\documentclass[final,5p,times,twocolumn]{elsarticle}

\usepackage{amssymb}
\usepackage{lipsum}
\usepackage{tabularx,booktabs,adjustbox}
\usepackage{amsmath}
\usepackage{xcolor}
\usepackage{epstopdf}
\DeclareGraphicsExtensions{.png,.jpg,.pdf,.eps}
\pdfoutput=1
\makeatother
\usepackage{soul}
\usepackage{hyperref}
\hypersetup{colorlinks,allcolors=green}
\usepackage{setspace}
\usepackage{amssymb}
\usepackage{xcolor}
\DeclareGraphicsExtensions{.png,.pdf}
\usepackage{longtable}

\journal{Advanced Engineering Informatics}

\begin{document}

\begin{frontmatter}



\title{Human Centric General Physical Intelligence for Agile Manufacturing Automation}


\author[label1]{Sandeep Kanta}

\author[label2]{Mehrdad Tavassoli}

\author[label3]{Varun Teja Chirkuri}

\author[label4]{Venkata Akhil Kumar}

\author[label5]{Santhi Bharath Punati}
\author[label4]{Praveen Damacharla}
\author[label7]{Sunny Katyara\corref{cor1}}

\affiliation[label1]{organization={Northeastern University}, 
addressline={354 Richards Hall, 360 Huntington Avenue},
  city={Boston}, 
  postcode={02115}, 
  state={Massachusetts}, 
  country={USA}}

\affiliation[label2]{organization={Consiglio Nazionale delle Ricerche}, 
  addressline={Corso Ferdinando Maria Perrone 24}, 
  city={Genova}, 
  postcode={16149}, 
  state={Liguria}, 
  country={Italy}}

\affiliation[label3]{organization={Columbia Sportswear}, 
  addressline={911 Southwest Broadway}, 
  city={Portland}, 
  postcode={97205}, 
  state={Oregon}, 
  country={USA}}

  \affiliation[label4]{organization={KINETICAI INC}, 
  addressline={1095 Evergreen Cir STE 433}, 
  city={The Woodlands}, 
  postcode={77380}, 
  state={Texas}, 
  country={USA}}

\affiliation[label5]{organization={Sunbelt Rentals Inc}, 
  city={Fort Mill}, 
  postcode={29715}, 
  state={South Carolina}, 
  country={USA}}

\affiliation[label7]{organization={Bmade Robotics}, 
  addressline={University College London}, 
  postcode={E152GW}, 
  country={UK}}

\cortext[cor1]{This research is supported by EI capital funding under 2030 Spin-off Initiatives (EINO: 25EI6716005). Correspondence Email: k.sunny@ucl.ac.uk}

\begin{abstract}
Agile human-centric manufacturing increasingly requires resilient robotic solutions that are capable of safe and productive interactions within unstructured environments of modern factories. While multi-modal sensor fusion provides comprehensive situational awareness yet robots must also contextualize their reasoning to achieve deep semantic understanding of complex scenes. Foundation model particularly Vision-Language-Action (VLA) models have emerged as promising approach on integrating diverse perceptual modalities and spatio-temporal reasoning abilities to ground physical actions to realize General Physical Intelligence (GPI) across various robotic embodiments. Although GPI has been conceptually discussed in literature but its pivotal role and practical deployment in agile manufacturing remain underexplored. To address this gap, this practical review systematically surveys recent advances in VLA models through the lens of GPI by offering comparative analysis of leading implementations and evaluating their industrial readiness via structured ablation study. The state of the art is organized into six thematic pillars including multisensory representation learning, sim2real transfer, planning and control, uncertainty and safety measures and benchmarking. Finally, the review highlights open challenges and future directions for integrating GPI into industrial ecosystems to align with the vision of Industry 5.0 for intelligent, adaptive and collaborative manufacturing ecosystem. 
\end{abstract}


\end{frontmatter}




\section{Introduction}

As global market pivots from mass production to hyper-personalized promise of Industry 5.0, a new industrial revolution is underway, one that prioritizes agility, human-centricity and customization over scalability and uniformity. Yet, for Small and Medium-sized Enterprises (SMEs) this transformation is more of aspirational than reality. Despite being the backbone of most economies, SMEs are often held back by limited technological infrastructure, tight budget and shortage of skilled workforce that make a leap to mass customization a formidable challenge. Recent studies across Europe and emerging economies reveal a starking truth that in some regions, as few as 3–5\% of SMEs have successfully embraced Industry 5.0 technologies that leaves vast majority at risk of falling behind in the race of future-ready manufacturing \cite{patalas2025mass}\cite{gold2025effects}.

Apart from logistical hurdles including high implementation costs, workforce training and supply chain complexities, the lack of reliable, standardized data and the intricate nature of integrating such Industry4.0 technologies with legacy industrial systems make deploying flexible robotic systems particularly challenging within existing industrial settings. These challenges are exacerbated by interoperability gaps, limited standardization in system architectures and need for modular and scalable solutions to facilitate integration and cost reduction \cite{patalas2025mass}\cite{katyara2024machine}. Although modern robots are equipped with advanced sensors and they operate on sophisticated machine learning (ML) architectures yet they still struggle to accommodate high degree of variability across industrial tasks, environments and embodiments. This is partly due to extensive task-specific retraining requirement and limited generalizability of conventional ML models that often require large and high-quality datasets with significant computational resources to adapt to new scenarios \cite{firoozi2025foundation}\cite{singh2024advances}.

    \begin{figure}[t]
       \centering
       \includegraphics[width=8.5cm]{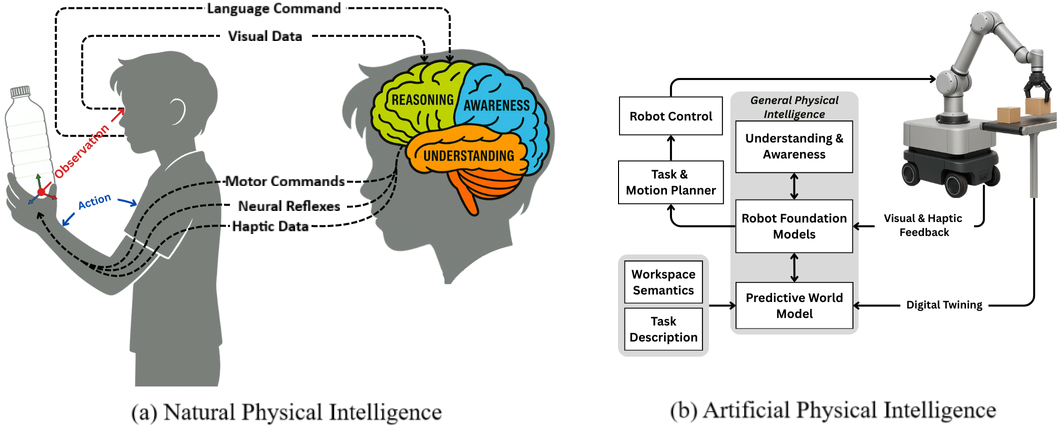}
       \caption{Conceptual representation of human-inspired general physical intelligence in the context of industrial automation.}
       \label{figure1}
       \vspace{-15pt}
    \end{figure}

To address these limitations, Foundation Models (FMs) have emerged as promising alternatives that leverage pretraining on massive and diverse corpus to develop broadly applicable models with enhanced transfer learning and zero-shot capabilities \cite{singh2024advances}. However, their performance is contingent upon the availability of large-scale, robot-relevant pretraining datasets and extensive real-world human demonstrations for fine-tuning both of which entail significant efforts to collect and substantial computational resources to engage. Additional challenges include ensuring safety, real-time performance and robust uncertainty quantification in stochastic industrial environments \cite{kawaharazuka2024real}. These factors currently limit the widespread adoption of FM-based systems in industrial applications. Ongoing research into simulation-based training, synthetic data generation and collaborative data-sharing initiatives are paving the way to overcome these barriers by enabling more scalable, adaptable and efficient deployment of agile robotic systems \cite{xiao2025robot}. 
 
Given the dynamic nature of robotic workspaces in agile manufacturing scenarios and the limitations inherent to FM implementations, an adaptable intelligence framework is exigent that reasons beyond what is explicitly presented to them and ranges to complex interpretations and unstructured commands within context. In this regard, modern generative AI frameworks based on Large Language Models (LLMs) including OpenAI GPT series and Google Gemini~\cite{team2025gemini} have offered great potential. Exemplified by recent multimodal and multi-task models including Gato~\cite{reed2022generalist}, PaLM-E~\cite{driess2023palm} and RT-2~\cite{zitkovich2023rt}, these models excel at integrating shared vision-language embedded representations to generate sequences of concrete control commands~\cite{lai2024vision}\cite{hu2023toward}. It is worth noting that however the efficacy of these models strongly depends upon rich sensory inputs such as vision and audition sensing as well as unambiguous user commands. The resultant contextual awareness itself is  prerequisite to a safe and dependable operation in human-robot shared industrial workspaces~\cite{bhatt2024know}\cite{oh2025towards} \footnote{According to neuroscientists, the primary function of the sepain brain is to generate adaptable and complex movements by integrating multisensory information from vision, audition and somatosensation.}. 

Regardless of their high-level reasoning, translating cognitive intent on “What to do” into actionable motor commands on “How to do” is not straightforward which is the primary reason that why true agile robotic systems have not been realized yet ~\cite{zafar2024exploring}. To facilitate better integration, recent implementation shifts away from modular architectures to more unified models that accounts for entire pipeline from perception and reasoning to action generation. This end-to-end design forms the central premise of what is known as GPI that effectively addresses the long-standing integration gap, as illustrated in Figure~\ref{figure1}\footnote{GPI enables robots to perceive (through visual and haptic feedback), reason (via foundation models and predictive world models) and act (through task planning and control) with semantic task understanding and context-aware awareness.}

The key enabling technology behind GPI are VLA models, classified as robotic FMs, that fuse perceptual inputs with natural language commands from users to form a multimodal representation that is shared across vision and language and grounded into physical actions to establish generalistic control framework. These models aim to drive synergy between precision of machines and nuanced human-like perception-action reasoning that make them particularly appealing for complex agile manufacturing takeovers~\cite{ma2024survey} beyond the capabilities of conventional pre-programmed control protocols~\cite{fan2024vision}\cite{wu2025h2r}. 

VLA models have demonstrated remarkable capabilities across multi-faced domains. For instance, MOSAIC~\cite{mishani2025mosaic} showcases their capacity on zero-shot learning thereby inferring object properties and planning manipulations without task-specific training and inference. Meanwhile, OrionNav and VLMap highlight their potential for spatial awareness by integrating multimodal embeddings with semantic mapping to achieve robust navigation~\cite{devarakonda2024orionnav}\cite{huang2022visual}. Models like Saycan~\cite{ahn2022can} translate natural language directives into feasible and structured plans that are grounded into low-level control commands\cite{silver2022pddl}\footnote{TAMP using PDDL is not merely a sequential task-then-motion pipeline nor is it restricted to fully observable or deterministic environments.} thereby bridging the gap between verbal instructions and high-level planning. Despite the progress they have achieved, they remain largely domain-specific and non-contextual. Thus, they fall short of capturing the broader perspective the GPI is trying to embody which is a generalistic and unified model capable of simultaneously learning and transferring knowledge throughout different interlinking sensory and cognitive modalities.

While VLA models have laid GPI foundation, they rely on linguo-visual information for the most part. In industrial setups however visual perception is often partial or occluded and thus cannot fully capture the dynamics of the scene. Therefore, supplementary sensing modalities—particularly haptic feedback and reflexes sensing must be integrated especially for contact-rich fine manipulation and slippage detection tasks.
Such complementary sensing gives rise to the concept of world models which play a crucial role in advancing Technology Readiness Levels (TRLs) not only through multimodal sensor integration but also via predictive simulation of future states. In doing so, these systems enhance their performance and reduce the risk, cost and iteration time associated with real–world experimentations~\cite{hofer2021sim2real}\cite{pashevich2019learning}. By encapsulating the underlying physical  dynamics governing the system–environment exchanges, the world models support robust decision–making and long–horizon planning~\cite{jauhri2022robot}\cite{zhu2021deep}. NVIDIA Cosmos has clearly exemplified such models by employing generative architectures to model high-fidelity environment dynamics for complex applications~\cite{agarwal2025cosmos} however its autoregressive predictions suffer from blurry outputs and are sensitive to conditional outputs. 

Moreover, a successful general–purpose model that facilitates robust reasoning and action generation in dynamic environments should incorporate main features including semantic understanding (from VLAs), physical sensation (from haptic interfaces) and predictive foresight (from world models). The convergence of these three pillars allows GPI to be intuitively guided through voice, haptic and reflexes instructions thereby fulfilling the core requirement for effective human–robot co-existence in agile manufacturing settings.

\begin{figure*}[t]
       \centering
       \includegraphics[width=18cm]{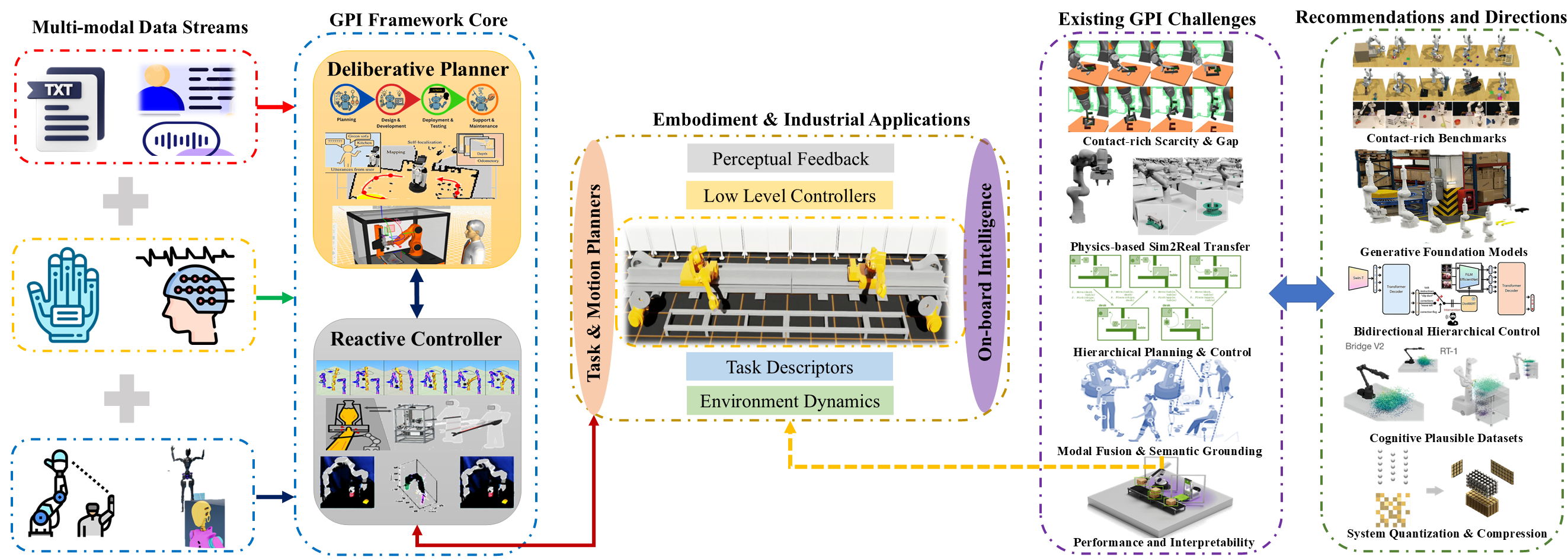}
       \caption{Graphical synopsis summarizing transformation of Vision-Language-Action models into General Physical Intelligence systems to advance agile manufacturing through multimodal perception and control for embodied industrial applications. Emphasis is on contact-rich interaction, transfer learning and semantic grounding with focus on development of robotics foundation models and cognitive datasets to enable safe and context-aware industrial robotic systems aligned with Industry 5.0. }
       \label{prime}
       \vspace{-15pt}
    \end{figure*}

Overall, GPI  aims to move beyond domain–specific learning by promoting a generalist model that integrates perception, reasoning and action while remaining physically grounded in real environmental conditions. Such adaptability is essential to realize human-centric doctrine of Industry 5.0 and motivates the research directions that are explored in this survey paper.
Thence, this paper (Figure~\ref{prime}) addresses following research questions:

\begin{itemize}
\setlength\itemsep{0.2em}
    \item To what extent existing VLA foundation models are technically viable to deliver GPI in agile manufacturing context especially for contact-rich interaction tasks?.
    \item What data foundations, multimodal fusion mechanisms and hierarchical control architectures are required to turn VLA models into physically grounded and safety-aware GPI systems?.
    \item How do different instantiations of GPI architecture built on top of existing VLA models including Gato-GPI, RT2-GPI, PaLM-E-GPI, OpenVLA-GPI could offer trade off on success rate, generalization, pose accuracy and task-level cycle time in a representative industrial benchmarks i.e., nult-bolt assembly and timber-cassette construction?
\end{itemize}

The rest of the paper is organized as;
\ref{section2} systematically reviews the foundational models underlying GPI and classifies them into six thematic keystones. Section \ref{section3} presents proposed GPI framework and validates its design through an ablation study that quantifies the contribution of each component to overall system performance. Section \ref{section4} discusses the practical challenges of real-world GPI deployment and outlines future research directions.
Finally, section \ref{section5} concludes this paper by summarizing the main findings and highlighting prospective avenues for industrial adoption.

\section{State-of-the-Art Frameworks}\label{section2}

Agile manufacturing environments are inherently dynamic and are characterized by high variability where close human-robot collaboration is frequent and inevitable. In this context, GPI role is crucial in enhancing flexibility and adaptability of automation systems. Such adaptability hinges on the robot system ability to synchronize perception, reasoning and physical actions in real time that are often guided by kinetic–dynamic models of workspace for rapid replanning while facing unforeseen events\footnote{Digital twins combined with kinetic–dynamic models support real-time task re-planning in agile and stochastic workspaces.}. Accordingly, this section examines recent progresses in GPI–enabled automation and structures them into six thematic pillars including;

\begin{enumerate}
    \item Multimodal 3D Representation, Grounding and Embodied Spatial Reasoning
    \item Data generation, simulation and sim-to-real transfer
    \item Long–horizon planning and skill sequencing
    \item Action generation and embodied control
    \item Uncertainty estimation and safety assurance
    \item Benchmarks and evaluation protocols
\end{enumerate}

\subsection{Multimodal 3D Representation, Grounding, and Embodied Spatial Reasoning}

Robust perception that is temporally aligned with motor actions is an indispensable part of GPI framework that facilitates context-aware decision making in variable environments. In practice, vision is primary exteroceptive modality in most perception units. While 2D spatial reasoning is of great importance for task execution and multimodal grounding \cite{you2023ferret} \cite{zhang2024ferret} but they cannot reliably infer 3D properties like metric depth, object size and true spatial relationships. Daxberger et al. \cite{daxberger2025mm} have addressed this issue in MM–Spatial that is specialist Multimodal Large Language Model (MLLM) trained on Cubify Anything VQA (CA–VQA) where RGB, depth and point clouds are jointly exploited for 3D-aware spatial reasoning and grounding. 
EmbodiedMAE \cite{dong2025embodiedmae} similarly targets embodement by jointly encoding multimodal visual information including RGB, depth and point clouds using multi-modal masked autoencoder trained based on large 3D robot manipulation dataset (DROID-3D) in order to obtain a unified 3D multi-modal vision backbone for embodied agents. Instead of collecting real 3D scenes, \cite{baik2025learning} learns 3D object–object spatial relationships (OOR) by prompting a pre-trained 2D image diffusion model that renders multi-view images of object pairs and lifting them to 3D using tri-angulation. These synthetic samples are then used to train text-conditioned score which is based on OOR diffusion model to generate new text-conditioned object layouts. GS–Reasoner  \cite{chen2025reasoning} together with the Grounded Chain–of–Thought (GCoT) dataset has gone one step further by fusing RGB vision patch features, point-cloud geometry and 3D positional cues into unitary token stream through dual-path pooling mechanism. Hence, it  enables direct autoregressive 3D visual grounding without increasing the number of tokens. To better align 3D semantic understanding with language instructions, SpatialReasoner \cite{liu2025neural} extends this concept to open-vocabulary 3D visual grounding with explicit spatial relations. In this accord, the LLM decomposes free-form queries into target, anchor and relations and then it queries visual property–enhanced hierarchical 3D feature field such as position, scale, color and CLIP features from SAM masks. To interpret implicit natural-language queries and adjust them accordingly to target object localization into 3D space across spatial, functional, logistic, emotional and safety reasoning, ReGround3D \cite{zhu2024scanreason} introduces 3D reasoning grounding and the ScanReason benchmark. It combines 3D-LLM reasoning module with geometry-enhanced 3D grounding module and a Chain-of-Grounding procedure to achieve state-of-the-art performance in 3D grounding and spatial reasoning.
While visual information is necessary, it is not sufficient on its own for full robot contextual awareness as robot vision is frequently compromised by noise, changing illumination, motion blur and occlusions. Therefore supplementary sensory information needs to be embedded into the latent shared representation. VLA models like Octo \cite{team2024octo} and Palm-E \cite{driess2023palm} account for these non-visual inputs. However, in Octo, initial experiments found that explicitly adding proprioceptive states sometimes hurt performance likely due to causal confusion between current robot state and future retrospective actions \cite{de2019causal}.

\subsection{Data generation, simulation and sim-to-real transfer}

The reliance of foundation models on large and diverse datasets presents bottleneck for robotics where collecting real-world data is both costly, time-consuming and non-dependable. Consequently, synthetically generated data, high-fidelity simulation and digital twins have become core contributors in the modern robot learning pipeline and thus to the development of GPI in agile manufacturing. In this regard, conventional sim-to-real pipelines are still largely built around domain randomization \cite{tobin2017domain} techniques where policies are exposed to randomized textures, lighting and physics parameters thereby improving their robustness towards visual and contextual shifts. However, selecting right parameter ranges is non-trivial and can lead to overly conservative policies if randomized too broadly and in non-Gaussian manner. To address these limitations, modern generative engines such as RoboGen \cite{wang2023robogen} and GenSim \cite{wang2023gensim} employ diffusion and autoregressive models to synthesize diverse, physics-consistent trajectories and scenes that significantly accelerates robot learning process. The idea is further expanded by large-scale simulators and knowledge bases e.g., RFUniverse \cite{fu2022rfuniverse} and ActivityPrograms \cite{nguyen2025regen}—which provide extensive libraries of objects and interactions. In the industrial context, such possibilities highlights high fidelity digital twins, in which the replicas of the cells and production lines are used to stress-test policies before deployment and to minimize  discrepancies between simulated and real setups \cite{long2025survey}, \cite{nazarczuk2025muble}, \cite{liu2025aligning}. Surveys on embodied AI and industrial cyber–physical systems further highlight the role of simulators (Isaac Sim, Habitat, iGibson, SAPIEN), automated scene generation (RoboGen, HOLODECK, ProcTHOR) and industrial metaverse / cognitive digital twins in generating multimodal data at scale and aligning virtual factories with their physical counterparts \cite{long2025survey}, \cite{lazaroiu2024cognitive}.
Beyond purely external simulators, learned world models provide an internal simulation foundation that supports imagination-based planning and data generation. DayDreamer\cite{wu2023daydreamer} exemplifies such efforts where latent dynamics model is trained jointly on simulated episodes and real-world robot trajectories, so that the agent can roll out candidate futures entirely in latent space and evaluate them before acting. By planning over these imagined rollouts and only executing the best action sequences on hardware, DayDreamer achieves successful transfer to real-world logistics and tabletop manipulation tasks while keeping real-world interaction data to a minimum \cite{wu2023daydreamer}. Regardless of their benefits, the fidelity of learned world models typically wears off over long prediction horizons thus limiting their effectiveness on highly compositional and multi-step industrial workflows. At the same time, cross-domain pretraining and self-improvement pipelines such as VC-1\cite{majumdar2023we} and RoboCat \cite{bousmalis2023robocat} aim to generalize across tasks and embodiments via large-scale pretraining and autonomous data collection at the cost of considerable compute and data requirements depending upon robot morphology and sensor configuration.
At the dataset level, a major milestone for data-driven sim-to-real is the Open X-Embodiment (OXE) dataset \cite{vuong2023open} which is a unified collection of trajectories from 22 robot embodiments that underpins generalist policies such as RT-X and Octo \cite{o2024open,team2024octo}. These models co-train on heterogeneous datasets (OXE, Bridge, DROID, ARIO, etc.) and demonstrate unprecedented cross-embodiment generalization.  However, recent evidence from CORTEXBENCH benchmarks shows that naively scaling data size and diversity mainly improves average performance across different embodiments and setups without any tangible domain-specific results \cite{majumdar2023we}. Majumdar et al. report that universal policies trained on OXE-style datasets still need substantial domain-specific fine-tuning to reach competitive performance in individual application regimes. Hence, universal compatibility across tasks remains contingent on further adaptation rather than being achieved purely through large-scale multi-embodiment pretraining \cite{majumdar2023we}.

Similar trends appear in large-scale VLA frameworks like RT-2 \cite{zitkovich2023rt}, $\pi_0$ \cite{black2024pi0} $\pi_{0.5}$ \cite{intelligence2025pi05}, GR00T N1 \cite{bjorck2025gr00t}, Gemini Robotics \cite{team2025gemini}, TriVLA \cite{liu2025trivla}, DiffusionVLA \cite{wen2024diffusion} and HAMSTER \cite{li2025hamster} which mix simulation benchmarks, real robot data, human videos and web-scale vision–language corpora to improve generalization and reduce per-cell data requirements. Yet, as reflected in Table \ref{tab:foundation_models}, many foundation policies remain biased toward vision and sparse kinematics like; RT-X models often lack unified coordinate frames and omit tactile and rich proprioceptive sensing which limits their applicability in contact-rich assembly, tight-tolerance insertion and force-sensitive operations \cite{o2024open,vuong2023open}. Clear criteria and standardized benchmarks for successful sim-to-real deployment in these regimes are still emerging.
A complementary line of work validates explicit sim-to-real transfer by first training controllers in simulation and then deploying them on real hardware with minimal adaptation.

In this regard, approaches such as SAM-E \cite{zhang2024sam}, MOSAIC \cite{mishani2025mosaic}, PLAN-SEQ-LEARN \cite{dalal2024plan} and DiffuseLoco \cite{huang2024diffuseloco} demonstrate explicit sim-to-real transfer wherein controllers are first trained in simulation and then deployed on real world setup with minimal domain adaptation efforts. These methods combine simulation-trained policies, skill libraries and diffusion-based controllers with limited real-world task data, domain randomization and local observations such as wrist-mounted cameras or orthographic 3D views to achieve robust transfer to real manipulators and legged platforms that are often in zero-shot or few-shot regimes.
Recent work further scrutinize real-to-sim-to-real video synthesis \cite{fang2025rebot}, chain-of-thought \cite{zawalski2024robotic} and chain-of-affordance \cite{li2025coa} data generation and automated robot-state–based annotation \cite{wen2025rosa} to scale up training data without manual labeling and to adapt pretrained VLA models to new domains with minimal in-domain samples.
In summary, the state of the art indicates a convergence toward integrated pipelines that combine (i) high-fidelity simulation and digital twins, (ii) generative data engines for scenes, trajectories and reasoning traces and (iii) large-scale cross-embodiment datasets and world models that all aimed at narrowing the domain gap in sim-to-real transfer for generalist robot policies in agile manufacturing environments. Meanwhile, accurately simulating complicated contact dynamics, friction, material deformation and human–robot interaction remains a major open challenge. Furthermore, long-horizon fidelity of learned world models and foundation policies under severe distribution shift is still insufficient for safety-critical agile production lines. These limitations motivate the subsequent discussion on long-horizon planning, embodied control and safety/uncertainty modeling.

\subsection{Long–horizon planning and skill sequencing}

Manufacturing tasks are intricate and consist of many steps where robots constantly need to infer human-defined instructions, adapt with varying elements in their surroundings and recover from erroneous configurations. In such circumstances, long-horizon planning and skill sequencing are undisputed qualities that needed to be manifested in GPI structure. Recent work spans hierarchical task planners, skill-centric frameworks and VLA-based policies that directly model extended action sequences.

A large bulk of work adopts hierarchical planning grounded into explicit skill libraries. This paradigm is instantiated by systems such as RePLan \cite{skreta2024replan}, MOSAIC \cite{mishani2025mosaic}, PLAN-SEQ-LEARN \cite{dalal2024plan}, Text2Motion \cite{lin2023text2motion}, STEP Planner \cite{zhou2025step}, Logic-Skill Programming (LSP) \cite{xue2024logic} and Learn-Gen-Plan \cite{hao2025learn}, where language or logic-based high-level planners are combined with low-level controllers. Abstract goals are then decomposed into subgoals or skill sequences e.g., grasp, push, insert that are geometrically and dynamically feasible. The main differences between these systems lie in the way they search over skill graphs,that can be via tree search, continuous optimization or hybrid search-and-shooting schemes. Yet they share common objective of composing generic skills into long-horizon manipulation plans.
To systematically evaluate such planners, simulation environments and benchmarks such as MuBlE \cite{nazarczuk2025muble} and BOSS \cite{yang2025boss} explicitly target long-horizon planning. MuBlE provides a MuJoCo–Blender–based simulation environment together with multi-step manipulation scenarios for evaluating task and skill planners, whereas BOSS exposes observation space shifts when skills are lined up, revealing considerable performance drop as observations drift across stages. In this accord, recent surveys on embodied and cyber–physical AI emphasize hierarchical planning, task scheduling and long-horizon coordination for complex industrial workflows \cite{long2025survey}, \cite{liu2025aligning}, \cite{ren2024embodied} while metaverse-oriented studies further highlight cooperative multi-agent control as a key mechanism for orchestrating such workflows in immersive industrial settings \cite{lazaroiu2024cognitive}.

Concurrently, VLA and foundation models increasingly internalize planning and skill sequencing. OneTwoVLA \cite{lin2025onetwovla}, RT-2 \cite{zitkovich2023rt} and Embodied Chain-of-Thought (E-CoT) \cite{zawalski2024robotic} explicitly use chain-of-thought or plan-then-act prompting to generate multi-step semantic plans before producing low-level actions to enable recovery from execution errors and flexible chaining of skills in tasks such as cooking or table-top assembly. CoA-VLA \cite{li2025coa} and DiffusionVLA \cite{wen2024diffusion} structure long-horizon reasoning through intermediate representations—chains of affordances or self-generated reasoning traces—that decompose complex tasks into explicit object, grasp, spatial and movement decisions or reusable reasoning segments for multi-task settings. Large-scale embodied agents such as Gemini Robotics~\cite{team2025gemini}, $\pi_{0.5}$~\cite{intelligence2025pi05} combine multimodal perception with memory and predictive modules to carry out extended activities such as; household cleaning, game-like tasks in novel environments, highlighting the role of internal state, memory and predictive dynamics for sustained skill sequencing.

At the interface between planning and control, recent policies predict temporally extended action sequences rather than single-step commands. SAM-E \cite{zhang2024sam}, CogACT~\cite{li2024cogact}, DexVLA~\cite{wen2025dexvla}, $\pi_0$~\cite{black2024pi0}, Octo~\cite{team2024octo}, DiffuseLoco \cite{huang2024diffuseloco} output action chunks or multi-step trajectories per inference that are often executed in a receding-horizon fashion. This design exploits temporal smoothness in manipulation and locomotion to enhance coherency and efficacy of executions for long-horizon tasks, for example laundry folding, factory sorting and multi-skill locomotion—while still allowing frequent replanning. Hierarchical architectures like HAMSTER \cite{li2025hamster} and TriVLA \cite{liu2025trivla} explicitly separate high-level VLM-based path or dynamics prediction from low-level control using intermediate 2D path or predictive state representations to facilitate long-horizon goal satisfaction and robust skill transitions in open-world settings. Diffusion and subgoal-based approaches such as SuSIE \cite{black2023zero} further decompose tasks into reachable visual subgoals to alternate between subgoal generation and low-level control to chain many instructions in both simulation and real environments.

Overall, the state of the art reveals rapid progress toward compositional and long-horizon behavior while highlighting several key challenges. Many systems lack guarantees on plan feasibility, safety or task completion under distribution shift and empirical studies highlight the accumulation of errors and observation drift when skills are chained. For agile manufacturing, an open research avenue is the tighter integration of these planning and skill-sequencing mechanisms with the multisensory representations to let the system properly accommodate the changes in its surroundings.

\subsection{Action Generation and Embodied Control}

At the final layer of GPI pipelines, rich perceptual and planning signals need to be converted into low-level motor commands in a safe, precise and responsive manner. The resultant system transforms the conventional modular actuation stacks into industrial cyber–physical systems—where different models handle perception, planning and control separately—to end-to-end VLA policies that directly map perception and language to continuous robot actions.
The first step in this regard is to ground action generated through demonstrations and teleoperation. Although large-scale simulation speeds up robot skill acquisition yet it cannot fully capture the complexity, variability and occasional failures that happen in real factories. To bridge this gap, the data from unstructured task instances in demonstration are used to build more robust policies that better reflect real operating conditions. Methods such as MimicPlay \cite{wang2023mimicplay} and MUTEX \cite{shah2023mutex} leverage these demonstrations to construct structured task graphs by abstracting reusable skill sequences that support cross-domain generalization with minimal retraining that is a key necessity for high-mix production lines. These graphs serve as discrete scaffolds for control. In this accord, given a new task instance, the system instantiates and re-parameterizes a sequence of known skills rather than learning a controller from scratch. Reinforcement-learning frameworks such as Adaptive Agent (AdA) \cite{team2023human} further modularize these structures into hierarchical policies that can be adapted across tasks and embodiments while loop-aware strategies like VOYAGER \cite{wang2023voyager} and Inner Monologue \cite{huang2022inner} refine behavior online through self-correction, environment feedback and explicit loop structures in the policy. Despite their flexibility, RL-based controllers typically suffer from poor sample efficiency and require careful reward shaping procedure particularly in sparse-reward contact-rich manipulation settings which limits their direct applicability to data and safety-constrained industrial environments.
From a control point of view, many of these systems can be categorized by how they incorporate sensory feedback when generating actions in Open-loop approaches including affordance-based planners \cite{fallon2015architecture} and semantic-scene-graph methods \cite{li2022embodied} which generate complete action sequences in a single forward pass based on an initial scene understanding. However, they are effective only when dynamics are predictable and disturbances are rare thus they miserably fail when objects move, humans intervene or sensors degrade. In contrast, closed-loop systems iteratively replan the procedure by integrating new sensory information at execution time that makes them more robust to unpredictable events such as part misplacement, occlusions or mid-task layout changes. This open-loop/closed-loop distinction is particularly important for downstream implementation where LLMs are increasingly used to translate high-level natural language instructions into executable policies or code via frameworks such as ProgPrompt \cite{singh2022progprompt} and Code-as-Policies \cite{liang2022code}. Recent hierarchical code-generation frameworks further elevate this paradigm by composing multi-step tasks from reusable subprograms while preserving low-level safety checks thereby reinforcing the link between strategic planning and low-level actuation \cite{chen2025code}, \cite{chen2024roboscript}. Nonetheless, LLM-based planners can be vulnerable to prompt ambiguity and hallucination. Henceforth to ensure that generated code is both safe and verifiable for physical execution remains a significant challenge in safety-critical manufacturing contexts. Simulation environments such as MuBlE \cite{nazarczuk2025muble} complement such efforts by providing closed-loop visual–action and control–physics loops to let agents learn and evaluate embodied control policies in realistic virtual manipulation scenarios before actual deployment.
To make the low-level control layer more explicit, skill-centric frameworks at the interface of planning and control are put into use. MOSAIC \cite{mishani2025mosaic}, PLAN-SEQ-LEARN \cite{dalal2024plan}, Text2Motion \cite{lin2023text2motion}, STEP Planner \cite{zhou2025step}, LSP \cite{xue2024logic} and Learn-Gen-Plan all assume a library of parameterized skills e.g., pick, place, push, insert, slide and focus on generating executable trajectories or sequencing these primitives in a way that respects geometric and physical constraints. In MOSAIC, generators produce candidate trajectories and world configurations by leveraging diffusion models and motion planning algorithms to diversify feasible actions. PLAN-SEQ-LEARN uses visual model-free RL (DRQ-v2) to learn low-level policies directly from images with local wrist-camera observations and stage termination conditions and safety observer for contact-rich tasks such as nut assembly. Text2Motion \cite{lin2023text2motion} optimizes parameters of a small set of manipulation primitives such as; Pick, Place, Pull, Push based on learned Q-functions and dynamics models and executes them in closed loop in both simulation and on real framework using Franka Panda robotic arm. STEP Planner \cite{zhou2025step} and LSP \cite{xue2024logic} both operate on a tree-structured hierarchy of subgoals where the root stands for the overall task and the leaves—also known as terminal nodes—correspond to the subgoals itself. Each leaf is explicitly designed to turn into a primitive action and the associated robust skill policies can handle uncertainties and external disturbances at contact points. Learn-Gen-Plan \cite{hao2025learn} further introduces a ChatGPT-based Skill Generator that synthesizes or adjusts skills from multimodal information by blending hand-crafted and learned controllers for complex assembly tasks. 

Collectively, these frameworks showcase a span from symbolic plans to executable actions although their reliance on predefined skill sets and task-specific training still limits generality of the system. Foundation-model however takes a different route by embedding action generation directly into VLA architectures to treat motor control as a first-class output modality alongside language and perception. Sequence imitation systems such as SAM-E \cite{zhang2024sam} predict multi-modal action vectors—6-DoF end-effector pose, discretized rotation, gripper state and collision permission—as temporally coherent sequences by using multi-channel heatmaps and classification heads to produce smooth keyframe-aware manipulation in both RLBench simulation and real Franka Panda setup.

In SAM-E, this sequential decoding with explicit collision-related signals yields more consistent and collision-aware trajectories in multi-step tasks. More broadly, generalist VLA controllers—including RT-2~\cite{zitkovich2023rt}, Octo~\cite{team2024octo}, Gemini Robotics~\cite{team2025gemini}, $\pi_0$~\cite{black2024pi0}, $\pi_{0.5}$~\cite{intelligence2025pi05}, DexVLA~\cite{wen2025dexvla}, DiffusionVLA~\cite{wen2024diffusion} and OG-VLA~\cite{singh2025og} extend this notion by employing diffusion, flow-matching or transformer-based policy heads to map multimodal embeddings into robot-executable actions whether as action tokens, short-horizon trajectories or low-level control chunks. Across these models jointly learning action sequences conditioned on vision–language context is associated with improved robustness and multi-step task completion even though the specific control rates, safety mechanisms and representations differ between architectures.

These architectures broadly fall into two categories based on their action representation strategies i.e., discrete-token policies and continuous-control models. For example, RT-2~\cite{zitkovich2023rt} and OpenVLA~\cite{kim2024openvla}  benefit from Internet-scale pretraining that discretizes 6-DoF end-effector motions and gripper commands into textual action tokens to enable shared vocabulary across language instructions, visual perception and action commands. Octo~\cite{team2024octo} on the other hand  adopts conditional diffusion action head that predicts continuous actions of either end-effector or joint positions that outperforms MSE and discretized heads in both zero-shot and finetuning settings and allow new action spaces to be added via lightweight adapters. Meanwhile, Gemini Robotics~\cite{team2025gemini} combines a cloud-based backbone with local action decoder to deliver smooth dexterous control across diverse bi-manual and industrial tasks. Flow-based models such as $\pi_0$~\cite{black2024pi0} introduce the concept of action expert by which flow matching is used to drive high-frequency multimodal action sequences routed via a mixture-of-experts transformer to facilitate robust and fluent control across multiple robot embodiments. $\pi_{0.5}$~\cite{intelligence2025pi05} extends this concept to open-world settings with end-to-end control from multimodal input in unstructured environment. Diffusion-based models like DexVLA~\cite{wen2025dexvla}, TinyVLA~\cite{wen2025tinyvla}, HybridVLA~\cite{liu2025hybridvla}, DiffusionVLA~\cite{wen2024diffusion} and GR00T N1~\cite{bjorck2025gr00t} scale it further to prove that expressive sequence generators are capable of supporting dexterous manipulation, bimanual coordination and agile locomotion directly from offline datasets with real-time deployment on edge hardware. These models alongside classical DRL and imitation-learning controllers highlight a clear shift from hand-crafted control stacks toward the end-to-end and data-driven motor policies that unify perception, reasoning and action~\cite{liu2025aligning}.
The interface between high-level reasoning and low-level control is yet another scheme that the action generation in GPI format. Systems such as OneTwoVLA~\cite{lin2025onetwovla} and JARVIS-1~\cite{wang2024jarvis} generate actions directly from most recent reasoning or high-level plans instead of decoupling planning and control into separate modules. This tight coupling minimizes the latency, avoids the possible inconsistencies between planned and executed actions and lets the controller reconsider its choices as new observations arise. Hierarchical architectures like HAMSTER~\cite{li2025hamster}, SuSIE~\cite{black2023zero}, CoA-VLA~\cite{singh2025og} and OG-VLA introduce intermediate representations on 2D paths, visual subgoals, chains of affordances or orthographic 3D canonical views so that low-level controllers can follow robust geometry-aware references while remaining resilient to errors in high-level planning. For instance, HAMSTER conditions low-level policy on predicted 2D paths overlaid on 3D perception and proprioception to allow the controller to focus on accurate trajectory tracking while tolerating moderate errors. SuSIE~\cite{black2023zero} uses a diffusion-based model to generate visual subgoals that are then reached by goal-conditioned low-level policy which is operating directly in the image space to improve precision over language-conditioned baselines. CoA-VLA~\cite{li2025coa} conditions action generation on visual–textual chains of affordances to enable policy to infer graspable regions, free spaces and obstacles before executing motion. OG-VLA~\cite{singh2025og} uses orthographic 3D projections to estimate the next end-effector pose by combining 3D-aware reasoning with robust action generation under changing camera and robot configurations.
Embodied chain-of-thought and state-alignment frameworks such as E-CoT~\cite{zawalski2024robotic} and ROSA~\cite{wen2025rosa} go one step further by linking reasoning tokens, robot state and control outputs. ECoT-trained VLAs first autoregressively generate reasoning tokens—plans, sub-tasks, movement primitives, object locations and then action tokens which grounds motor commands into an explicit inspectable chain of reasoning to improve success rates on out-of-distribution tasks instances. ROSA uses robot state estimation to align high-level semantic instructions with low-level 3D action spaces to address spatial and temporal gaps between vision–language backbones and physical actuation and lead to more accurate and reliable control in dynamic environments. These approaches not only enhance performance but also improve interpretability and facilitate human oversight which is crucial in safety-critical manufacturing.
Optimization-oriented methods refine action generation to improve efficiency and reliability in real-world deployment. The OFT recipe in Fine-Tuning VLA~\cite{kim2025fine} enhance control throughput and smoothness via parallel decoding, action chunking, continuous action representations and an L1 regression objective to generate a 26 times increase in action-generation output rate and large gains in success rate in both simulation and on real bimanual ALOHA hardware~\cite{fu2024mobile}. Large diffusion-policy transformers such as DiffuseLoco~\cite{huang2024diffuseloco} demonstrate that diffusion-based controllers can generate stable and agile locomotion from offline datasets with transformer-based diffusion models producing low-level joint commands at real-time rates on edge devices while outperforming RL and non-diffusion baselines in stability and velocity tracking. Data-centric pipelines such as ReBot~\cite{fang2025rebot} improve robustness and out-of-domain generalization by augmenting training with real-to-sim-to-real video synthesis. In this case, real robot trajectories are replayed in simulation, objects and environments are diversified and the resulting motions are re-projected into realistic videos, used to train VLA policies~\cite{fang2025rebot}.

Finally, in-context adaptation frameworks like RICL~\cite{sridhar2025ricl} allow pre-trained VLA models to acquire new skills from only 10–20 demonstrations for manipulation  and grasping tasks without any model parameter updates. In this way, it retrieves only the relevant segments of the demonstrations into the model context which facilitates  in-context learning of new tasks and significantly boost performance in real-world manipulation scenarios~\cite{sridhar2025ricl}. This in-context control paradigm is particularly attractive for agile manufacturing where rapid changeover to new parts and fixtures is essential and retraining is often impractical.

It is worth noting that regardless of progress that has been made, there exist significant gaps between current action-generation architectures and what is required for a resilient and human-centric manufacturing cell. Imitation-based controllers and open-loop skill chains remain sensitive to shifts in observation space especially when skills are executed sequentially under changing visual conditions thus the observation-space drift leads to substantial drops in task success ~\cite{yang2025boss}. Taken together, these limitations extend across control paradigms i.e., RL-based controllers suffer from data inefficiency and reward-design issues, LLM-based code planners are vulnerable to hallucinations and lack rigorous verification and safety and VLA-based policies typically lack explicit reasoning over safety, uncertainty and hard physical constraints.

\subsection{Uncertainty estimation and safety assurance}

Safe and reliable operation is a non-negotiable priority in manufacturing environments where robots must remain robust to uncertainties and unplanned events while sharing workspaces with humans. Modern approaches to safe GPI therefore treat safety and uncertainty as multi-level concerns spanning planning, action and execution. At the architectural level, systems such as Physical AI Agents integrate anomaly detection, access to operational protocols and regulatory compliance into the decision loop and explicitly call for adaptive decision-making under uncertainty and multi-agent safety coordination in future platforms~\cite{long2025survey}. Complementary risk-aware controllers enforce safety margins, collision envelopes or fail-recovery protocols in shared workspaces to provide a classical control layer beneath cognitive policies~\cite{khan2025safety}.

A foundational element of safety is quantifying model confidence before acting. KNOWNO~\cite{ren2023robots} integrates conformal prediction to estimate planner confidence where if confidence falls below a threshold, the system triggers a safe stop or requests human intervention rather than executing low-trust plans. Text2Motion~\cite{lin2023text2motion} uses Q-function ensembles to detect out-of-distribution skills and rejects candidates whose Q-value variance exceeds a calibrated threshold to prevent symbolically or geometrically invalid actions that often lead to control exceptions. PLAN-SEQ-LEARN~\cite{dalal2024plan} evaluates robustness to noisy pose estimation and relies on stage-termination conditions and point-cloud-based collision checking as progress and safety monitors to reduce cascading failures across long-horizon tasks. Accurately calibrating confidence thresholds and ensuring that such uncertainty measures generalize across tasks and environments remains an open problem.

Uncertainty is also addressed at the policy level via probabilistic and diffusion-based control. Instead of producing a single deterministic action diffusion policies such as SUDD~\cite{ha2023scaling} and RDT-1B~\cite{liu2024rdt} model full distributions over actions to yield more calibrated decisions in high-dimensional control spaces typical of complex assembly and material-handling tasks. DiffuseLoco~\cite{huang2024diffuseloco} similarly uses a diffusion-based locomotion policy that reasons over movement affordances and collision-free trajectories to achieve smooth, stable and robust real-time control on legged robots. These approaches improve robustness under disturbances and modeling errors but introduce higher computational cost and potentially slower inference due to sampling which is problematic for high-frequency industrial controls.

Several frameworks embed explicit safety checks and semantic alignment into planning and execution. RePLan~\cite{skreta2024replan} employs a Verifier module to ensure that each plan step is necessary and consistent with its reward function by pruning unsafe or irrelevant actions and augments controller residuals with safety terms during real-robot deployment. Gemini Robotics~\cite{team2025gemini} combines content safety inherited from underlying Gemini model with robotics-specific safety post-training and evaluation on benchmarks such as ASIMOV~\cite{sermanet2025generating} while interfacing with classical safety-critical controllers for collision avoidance and force control. Language-conditioned control schemes such as RT-H~\cite{belkhale2024rt} use natural-language motion waypoints to backtrack and revise actions when execution diverges from the plan and mixture-of-experts controllers like $\pi_0$~\cite{black2024pi0} dynamically select among specialized policies to avoid unsafe behaviors in novel or challenging scenarios.

Finally, simulation-based risk mitigation aims to foresee and avoid unsafe behaviors before execution. Planning frameworks such as LLM-MCTS~\cite{zhao2023large} and RAP~\cite{kagaya2024rap} simulate candidate action sequences with world-model roll-outs while using LLM reasoning to filter high-risk or ineffective plans which is a capability that is crucial for certifying robots in human-interactive and tightly constrained environments. However, the effectiveness of these methods critically depends upon the fidelity and calibration of underlying world models that still struggle with rare events, complex contacts and non-stationary factory conditions. Overall, current work provides important building blocks—confidence estimation, OOD detection, probabilistic policies, verification modules and semantic safety layers but safety and uncertainty are still largely treated as add-on filters rather than first-class design principles tightly integrated with multisensory representation, sim-to-real pipelines and VLA/world-model architectures.

\subsection{Benchmarks and evaluation protocols}

To ensure industrial readiness, the robotic systems must be systematically evaluated to track progress, identify generalization gaps and validate their safety and efficiency. While invaluable, the performance in simulation does not always translate directly to the physical world due to persistent sim2real gap especially in contact dynamics and stochastic physics distribution. This requires multi-layered approach to benchmarking that spans from controlled simulations to real-world physical tests by using metrics that capture not only task success but also operational value. Initial evaluation often occurs in simulation due to its safety, low cost and scalability. Platforms like VIMA-Bench \cite{li2023mastering}, Ravens \cite{muppidisynthetic} and RLBench \cite{james2020rlbench} are invaluable for assessing model ability to generalize across structured tasks with controlled variability. These environments facilitate precise measurement of model capabilities in manipulation, tool use and object rearrangement when faced with novel objects and scenes. For agile manufacturing, this is crucial for cost-effectively testing on how system might adapt to new product designs or component variations before physical deployment and roll-outs. To meet the generalization challenges posed by these benchmarks, a two-stage fine-tuning strategy is often employed where models are pre-trained on simpler tasks and refined on more complex ones incrementally \cite{gao2023two}.

While simulation is powerful but performance in the physical world is the ultimate test. Real-world benchmarks like CALVIN \cite{mees2022calvin}, ALOHA \cite{fu2024mobile} and large-scale BridgeData V2 \cite{walke2023bridgedata} provide diverse tasks in physical environments with real sensors and actuators. These platforms are essential for evaluating system robustness to real-world noise, sensor delays and unmodeled physics. The ManiSkill2 \cite{gu2023maniskill2} benchmark further bridges this gap by providing broad suite of manipulation tasks that can be consistently compared across both simulation and reality thus offering more complete picture of model capabilities. However, conducting extensive real-world benchmarking is expensive, time-consuming and raises safety concerns. For agile manufacturing, the evaluation must extend beyond task completion rates. The practical readiness of a GPI-enabled platform is measured by KPIs that reflect industrial value such as cycle time, generalization score to new tasks and overall performance index \cite{hakimi2024key}\footnote{Cycle Time which is the average time required to successfully complete given manufacturing task. This is primary driver of throughput and operational efficiency in factory settings that directly impact production capacity and cost. Whereas, generalization score quantifies performance consistency across variations in task parameters (e.g., object position, material properties and lighting conditions). High generalization is crucial for agile manufacturing where production lines need to handle high-mix and low-volume products runs without costly and time-consuming reprogramming or retraining. And force-profile adherence evaluates how closely robot measured force-torque interaction profile matches expert-demonstrated profile during contact-rich tasks. This is vital for quality control in operations like fastening to ensure correct torque, polishing to maintain consistent pressure or delicate assembly where applying correct force dynamics is as important as completing the task itself.}. Standardizing these KPIs across different hardware platforms and task domains remains a challenge. Furthermore, as robots become more intelligent, new metrics are emerging to evaluate their cognitive alignment and interaction quality. Frameworks like Inner Monologue \cite{huang2022inner} and E-CoT \cite{zawalski2024robotic} introduce metrics for explainability, language grounding and iterative correction. These cognitive benchmarks are building trust and ensuring seamless coordination in human-robot teams. They are shaping evaluation of next-generation VLA systems not just by what they do but by how and why they do it as crucial step toward creating truly intelligent and dependable industrial partners. Quantifying these cognitive aspects objectively and reliably is difficult often relying on qualitative assessments or proxy metrics. 

Furthermore, as robots become more intelligent, new metrics are emerging to evaluate their cognitive alignment and interaction quality. Frameworks like Inner Monologue and E-CoT introduce metrics for explainability, language grounding and iterative correction. These cognitive benchmarks are essential for building trust and ensuring seamless coordination in human-robot teams. They are shaping the evaluation of next-generation VLA systems not just by what they do but by how and why they do it to represent crucial step toward creating truly intelligent and dependable industrial partners.

\begin{table*}[htbp]
\centering
\caption{Comparison of Vision-Language-Action Foundation Models for Robotics}
\label{tab:foundation_models}
\scriptsize
\begin{adjustbox}{width=\textwidth}
\renewcommand{\arraystretch}{1.1}
\begin{tabularx}{\textwidth}{
@{}
>{\raggedright\arraybackslash}p{1.3cm}
>{\raggedright\arraybackslash}p{2.5cm}
>{\raggedright\arraybackslash}p{3.0cm}
>{\raggedright\arraybackslash}X
>{\raggedright\arraybackslash}X
>{\raggedright\arraybackslash}p{2.8cm}
@{}}
\toprule
\textbf{Framework} & \textbf{Robotics Domain} & \textbf{Foundation Model} & \textbf{Characteristics} & \textbf{Limitations} & \textbf{Exploitation} \\
\midrule

OpenVLA~\cite{kim2024openvla} & Context-aware In-hand Manipulations & LLaMA 2~\cite{touvron2023llama} + DINOv2~\cite{oquab2023dinov2} + SigLIP~\cite{tschannen2025siglip} & 
7D control via discrete tokenization; generalist transformer policy (7B); dual visual encoders; strong multi-robot task generalization; LoRA-efficient fine-tuning and quantization & 
Single-image instance; no proprioception and temporal input; low inference rate (not suitable for 1 KHz); success rate capped at 90\% & 
970k robot demos from Open-X; evaluated on BridgeData V2 and Franka-DROID robots \\

\midrule
Cosmos~\cite{agarwal2025cosmos} & Physical Intelligence Agents & Diffusion-7B/14B and AutoReg-4B/12B & 
Multi-view video generation with spatial consistency; 3D-aware and physics-consistent outputs & 
Blurry outputs from auto-regressive variants; sensitive to conditioning type and model size & 
Internal video corpus; RealEstate10K, Bridge, Cosmos-1X, RDS \\

\midrule
RT-1~\cite{brohan2022rt} & Fine Manipulations & Custom Transformer + FiLM-efficientNet~\cite{tan2019efficientnet} & 
Learns 700+ tasks with 97\% seen-task accuracy; generalizes to unseen tasks and environments; real-time 1 kHz control & 
Imitation only without RL; limited generalization to novel motor behaviors; constrained to kitchen environments & 
130k real-world trajectories from 13 robots across 17 months \\

\midrule
RT-2~\cite{zitkovich2023rt} & Generalized Dexterous Manipulations & PaLI-X~\cite{chen2023pali} (5B, 55B) + PaLM-E~\cite{driess2023palm} (12B) & 
Combines web-scale vision-language data with real robot demonstrations; symbolic reasoning, multilinguality, chain-of-thought planning; improves generalization to unseen tasks and instructions via co-fine-tuning & 
Cannot generalize to unseen motor skills; relies on high-capacity VLMs; not suitable for high-frequency control (1 kHz) & 
Over 1M real-world trajectories from 13–22 robots; datasets include RT-X, Cable Routing, Jaco Play, RoboTurk, NYU VINN and Language Tables \\

\midrule
RT-X~\cite{o2024open} & Dexterous Manipulation & RT-1-X~\cite{o2024open}: FiLM-conditioned efficientNet~\cite{tan2019efficientnet} + Transformer (35M); RT-2-X~\cite{o2024open}: PaLI-X~\cite{chen2023pali} + UL2~\cite{tay2022ul2} VLA model & 
Transformer-based tokenized control policies; strong cross-embodiment generalization; RT-2-X handles unseen instructions and spatial relations; co-training improves small-data performance & 
RT-1-X underfits large-scale data; lacks unified coordinate frames; excludes tactile and proprioception sensing; no established transfer success criteria & 
Open X-Embodiment Dataset: 1M+ trajectories, 22 robots, 527 skills, 60 datasets across 21 institutions \\
\midrule
OK-Robot~\cite{liu2024ok} & Mobile Manipulation & CLIP~\cite{nguyen2025robotic}, OWL-ViT~\cite{minderer2022simple}, Lang-SAM, AnyGrasp~\cite{fang2023anygrasp} & 
Zero-shot manipulation using pretrained open-knowledge models; fast deployment with minimal scans; modular vision-language pipeline (detection, segmentation, grasping) & 
No task-specific training; static semantic maps; fails on ambiguous language; no recovery if one module fails & 
Evaluated in real homes with minimal prior data; heuristic-based object and grasp detection \\
\midrule
ChatGPT for Robotics~\cite{vemprala2024chatgpt} & 
Interaction Control & 
ChatGPT (GPT-3.5/4), CLIP~\cite{nguyen2025robotic}, PaLM-E~\cite{driess2023palm}, ROS bridge & 
Explores how ChatGPT can be integrated with robotic systems via modular planning and control; enables robots to interpret natural language commands and generate high-level task plans; interfaces with perception modules (e.g., CLIP) and robot APIs (e.g., ROS) for real-world execution; emphasizes modularity, prompt engineering and tool use for embodied reasoning & 
Vulnerable to prompt ambiguity and hallucination; real-time control and safety not fully addressed; performance depends on quality of connected perception and control modules & 
Evaluated qualitatively across diverse tasks (e.g., object retrieval, navigation); shows promise for general-purpose robotic interfaces via LLM prompting and reasoning \\
\midrule
PaLM-E~\cite{driess2023palm} & 
Adapative Robotic Control & 
PaLM~\cite{kim2024palm}, ViT~\cite{dosovitskiy2020image}, RT-1~\cite{brohan2022rt}, CLIP~\cite{nguyen2025robotic} embeddings & 
Extends PaLM with embodied multimodal inputs (images, robot states, language); supports planning and control via language-only prompts; integrates visual and proprioceptive data with language using a unified transformer; & 
Requires extensive compute and data for training with high inference latency; action space handled via pre-trained controllers (e.g., RT-1) instead of direct motor control & 
Benchmarked on tasks like SayCan, RT-1 and reasoning over visual inputs; demonstrates strong zero-shot generalization diverse over mobile manipulation tasks \\

\midrule
CLIPORT~\cite{shridhar2022cliport} & 
In-hand manipulation & 
CLIP~\cite{nguyen2025robotic}, Transporter & 
Two-stream architecture combining semantic (CLIP) and spatial (Transporter) pathways; enables end-to-end imitation learning from language commands; action-centric pick and place affordance prediction with frozen CLIP backbone & 
Struggles with long-horizon tasks; does not support dexterous 6-DOF control without memory or symbolic representations; can't predict task completion or handle partially observable environments & 
Evaluated on 10 simulated and 9 real-world tasks; benchmarked on Ravens environment with seen/unseen attributes \\
\midrule
VIMA~\cite{jiang2022vima} & Dexterous manipulation & Transformer with multimodal prompting (language + 3D vision) & 
Uses multimodal token sequences to condition behavior; demonstrates one-shot generalization with varying object combination & 
Sim-only training and evaluation limited proprioceptive or tactile input; high reliance on synthetic data and controlled prompt structure & 
20K+ procedurally generated tasks in Pybullet simulation; language-visual goal specifications \\

\bottomrule
\end{tabularx}
\end{adjustbox}
\end{table*}

\begin{table*}[htbp]
\centering
\caption{Comparison of Vision-Language-Action Foundation Models for Robotics (Continued)}
\scriptsize
\begin{adjustbox}{width=\textwidth}
\renewcommand{\arraystretch}{1.2}
\begin{tabularx}{\textwidth}{@{}p{2.2cm} p{2.6cm} p{2.5cm} X X p{2.8cm}@{}}
\toprule
\textbf{Framework} & \textbf{Robotics Domain} & \textbf{Foundation Model} & \textbf{Characteristics} & \textbf{Limitations} & \textbf{Exploitation} \\
\midrule

\midrule
SayCan~\cite{ahn2022can} & Task and Motion Planning & FLAN-T5 ~\cite{wei2021finetuned} (Language) + Skill affordance model (Reward predictor) & 
Grounds LLMs using affordance scores from real-world skill executors; decomposes tasks into steps and ranks executable actions based on predicted success & 
Fixed skill repertoire; limited generalization to new tasks without retraining; relies on pre-trained LLMs without end-to-end learning; LLM outputs not conditioned on environment state & 
Evaluated on 101 real robot tasks; demonstrations from the PaLM model combined with real-world execution on everyday household tasks \\

\midrule
EmbodiedGPT~\cite{mu2023embodiedgpt} & Human Robot Interaction & 
ViT~\cite{dosovitskiy2020image}, GPT-style decoder, CLIP~\cite{nguyen2025robotic}, BLIP-2~\cite{li2023blip}, LLaMA~\cite{touvron2023llama} & 
Proposes Embodied Chain of Thought (E-CoT) for reasoning-enhanced vision-language pretraining; pretrained on embodied trajectories with textual rationales and multimodals & 
Relies on synthetic embodied datasets; inference latency due to autoregressive token generation; challenges in long-horizon and high-frequency control; assumes availability of detailed language descriptions & 
Evaluated on ALFRED and TEACh benchmarks; shows improvement over baselines in language-conditioned tasks; ablations confirm importance of E-CoT and multimodal fusion \\

\midrule
Gato~\cite{reed2022generalist} & Multi-modal Iterative Control & Token-based, decoder-only Transformer & 
Unified Transformer trained via token-based imitation learning on vision, language, robotics and gaming; supports multi-task and multi-embodiment generalization with task prefixing & 
Lacks grounding in real-world robot embodiment and sensors; trades off per-task performance for generality; no closed-loop planning; high compute prevents real-time deployment & 
604-task corpus with real-world robot teleoperation, NLP corpora and environments like Atari and Minecraft \\

\midrule
GR00T N1~\cite{bjorck2025gr00t} & Anthropomorphic Grasping and Manipulation  & Dual-system VLA: Frozen Eagle-2~\cite{li2025eagle} + SmolLM2~\cite{allal2025smollm2} for VLM; Diffusion Transformer for control (DiT) & 
Supports bimanual coordination and inter-hand transfer; trained via cross-attention and flow-matching loss; modular embodiment encoding (LeRobot++) & 
Limited to short-horizon tabletop tasks;physics-inconsistent video generations reduce realism; post-training fine-tuning may overwrite emergent behavior & 
Multi-source training: Ego4D, Epic-Kitchens, Assembly101; 780K sim demos (DexMimicGen); real-world GR-1 teleoperation \\

\midrule
RoboCat~\cite{bousmalis2023robocat} & Task Agonistic in-hand Manipulation & 
ViT~\cite{dosovitskiy2020image}, autoregressive transformer, Action-Tokens, self-supervised fine-tuning & 
Trains a single model across multiple robots and tasks; supports self-improvement through fine-tuning on collected data; enables cross-embodiment and cross-task generalization; & 
Action representation abstracted from low-level control; high data and compute requirements for self-improvement loops; performance varies with embodiment & 
Trained on 141k demonstrations across 21 tasks and 4 robot embodiments; evaluated on real robots with up to 50\% success rate on unseen tasks after fine-tuning \\
\midrule
RoboFlamingo~\cite{li2023vision} & 
Generalized Learning from Demonstrations & 
Flamingo~\cite{alayrac2022flamingo}, ViT~\cite{dosovitskiy2020image}, CLIP~\cite{nguyen2025robotic}, Flamingo-robot adapter & 
Frozen backbone ensures stability while training adapter; shows strong generalization from internet-scale pretraining to embodied robotic; capable of goal-conditioned control from images and language & 
Computationally intensive and relies on adapter tuning and high-quality demonstrations; no active planning or feedback integration; evaluated mainly on imitation with no reinforcement or planning-based settings & 
Benchmarked on 34 tasks across 3 real-world robot arms; tested on new robot embodiments and unseen tasks; shows improved performance with few-shot samples with frozen encoders and trained adapters \\

\midrule
RT-H~\cite{belkhale2024rt} & 
Task and Motion Planning & 
RT-2~\cite{zitkovich2023rt}, PaLM~\cite{kim2024palm}, ViT~\cite{dosovitskiy2020image}, RT-H policy~\cite{belkhale2024rt} & 
Extends RT-2 with hierarchical control to decompose high-level instructions into low-level actions using vision-language planning policies; enhances generalization to long-horizons with minimal tuning & 
Depends on quality of subgoal decomposition and subgoals are generated offline and are not updated dynamically during execution; not end-to-end trainable and limited real-world testing on diverse robotic platforms & 
Demonstrates improved success on multi-step instructions over RT-2; benchmarked on task composability and generalization with language-derived action hierarchies \\
\midrule
$\pi_0$~\cite{black2024pi0} & Robust Grasping and Fine Manipulation & ViT~\cite{dosovitskiy2020image}, CLIP~\cite{nguyen2025robotic}, TokenLearner, GPT-style architecture & 
Introduces $\pi_0$ transformer that maps vision and language to actions; enables consistent performance across multiple embodiments and tasks; lightweight and efficient due to token pruning & 
Trained with frozen vision encoders; lacks fine-grained 3D spatial reasoning; limited real-world evaluation with rapid feedback controls & 
Evaluated on $250+$ tasks across 6 robot embodiments; benchmarked on RLBench, BridgeData v2 and Open X-Embodiment datasets \\

\midrule
RT-Trajectory & 
Dexterous Manipulation Planning & 
RT-1~\cite{brohan2022rt}, sketch encoder, trajectory-conditioned policy, CLIP~\cite{nguyen2025robotic} embeddings & 
Uses hindsight trajectory sketches for task abstraction and sketch-conditioning for task generalization; improves compositional generalization by aligning visual observations with trajectory sketches& 
Depends on diversity of sketch annotations; no explicit hierarchical planning or long-horizon memory; real-world deployment is limited to table-top environments & 
Evaluated on 18 real-world manipulation tasks; shows improved generalization to unseen tasks compared to RT-1; validates benefits of trajectory sketch conditioning for policy learning \\

\midrule
Act (Actor) & 
Dexterous Bimanual Manipulation & 
Transformer-based policy (Act), ViT~\cite{dosovitskiy2020image}, GPT-style decoder, RGB and proprioception inputs & 
Operates with RGB cameras and proprioception without depth; efficient and scalable policy learning without expensive hardware & 
Limited to short horizon tabletop tasks; sensitive to lighting and viewpoint variations due to RGB-only input & 
Evaluated on 10 real-world bimanual tasks; achieves high success rates with fewer than 500 demonstrations per task \\

\bottomrule
\end{tabularx}
\end{adjustbox}
\end{table*}

\section{General Physical Intelligence Framework and Ablation Study} \label{section3}

   \begin{figure}[t]
      \centering
      \includegraphics[width=8.5cm]{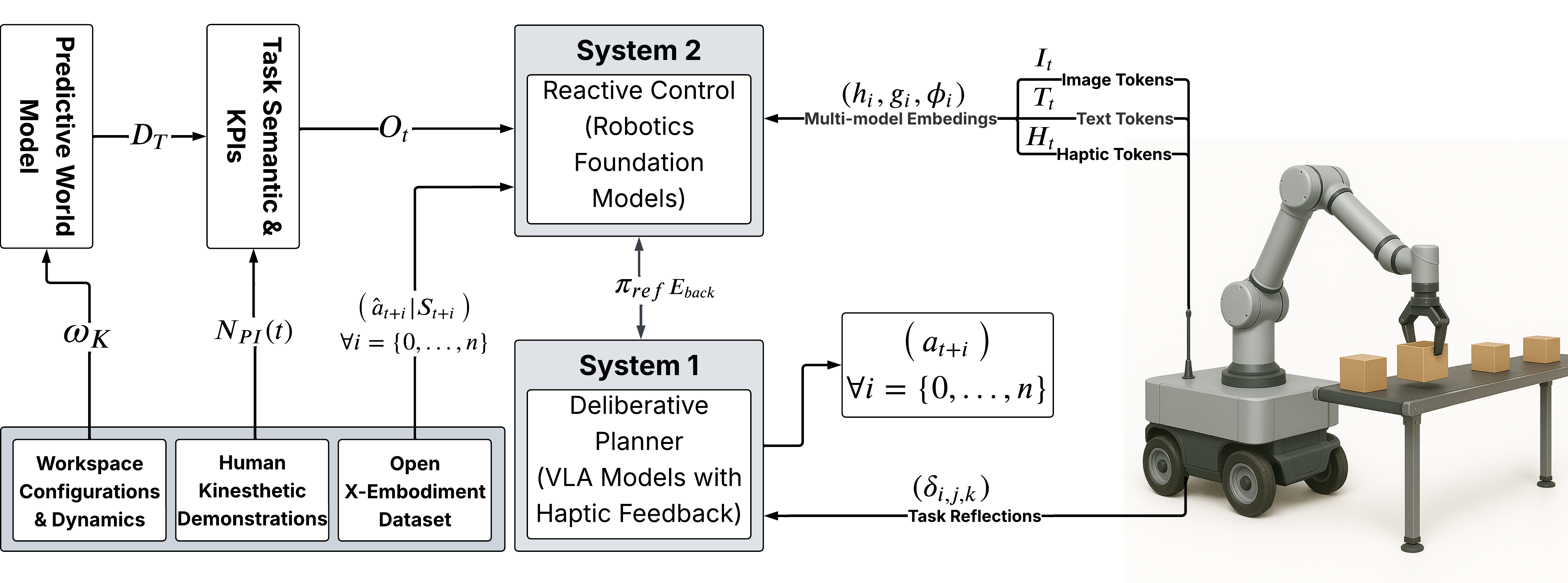}
      \caption{Architectural representation of general physical intelligence in the context of agile manufacturing.}
      \label{figure2}
      \vspace{-15pt}
   \end{figure}

General Physical Intelligence (GPI) is ability of robotic agent to perceive, reason and act upon physical world by mapping multimodal sensory inputs including vision, language, proprioception, haptics, machine state to physically grounded actions across diverse embodiments and tasks. This requires not only advanced data-driven models but also explicit integration of physical laws and models for robust, explainable and generalizable behaviour \cite{Mengaldo2022}\cite{Zhao2025}. The GPI policy can be formalized as [ $a_t \sim \pi_\theta(a_t \mid \tau(o_{\leq t}), \hat{s}_{t:t+H}, \mathcal{G}, \mathcal{E})$] where $\tau(o_{\leq t})$ is the tokenized history of multimodal observations, $\hat{s}_{t:t+H}$ is the predicted sequence of future states from a world model, $\mathcal{G}$ is the task or goal specification and $\mathcal{E}$ is the embodiment descriptor e.g., kinematics, actuation limits. 

Suppose a robot uses a force/torque sensor to grasp an object. The relationship between the applied force $F$ and the resulting displacement $x$ in a compliant gripper finger can be modeled by Hooke’s Law with $F = -k x$ where $k$ is the stiffness constant of gripper. This law is embedded in the robot world model and used to predict the outcome of actions involving elastic deformation such as adaptive grasping and compliant assembly.

Learning objective for GPI agent incorporates action prediction loss ($\mathcal{L}_{\text{action}}$), forward model prediction loss ($\mathcal{L}_{\text{world-model}}$) and task success ($\mathcal{L}_{\text{goal-success}}$) governed by physics-informed penalties to enforce compliance with known physical laws as seen in recent physics-informed machine learning (PIML) and physics-informed neural network (PINN) approaches as defined using Eq. \ref{eqt2} \cite{alonso2025mathematics}

\begin{equation}
    \mathcal{L}_{\text{GPI}} = \mathcal{L}_{\text{action}} + \lambda_1 \mathcal{L}_{\text{world-model}} + \lambda_2 \mathcal{L}_{\text{goal-success}}
\label{eqt2}
\end{equation}

\noindent where:
\begin{align*}
    \mathcal{L}_{\text{action}} &= -\sum_{t} \log \pi_\theta(a_t \mid \cdot) \\
    \mathcal{L}_{\text{world-model}} &= \sum_{t=1}^{T} \left\| \hat{s}_{t+1} - s_{t+1}^{\text{true}} \right\|^2 \\
    \mathcal{L}_{\text{goal-success}} &= - \mathbb{E}_{\tau} \left[ R(\tau, \mathcal{G}) \right]
\end{align*}

$R(\tau, \mathcal{G})$ is cumulative reward function for achieving task conditioned upon goal $\mathcal{G}$ with $\lambda_1$ and $\lambda_2$ be weight coefficients.

Leading robotics foundation models such as Gato \cite{reed2022generalist}, RT-2 \cite{zitkovich2023rt}, PaLM-E \cite{driess2023palm}, OpenVLA \cite{kim2024openvla} and VIMA \cite{jiang2022vima} provide basis for GPI but require significant extensions to be effective in agile manufacturing. This involves more than just adding sensory inputs as it requires architectural and training modifications to build a deeper awareness of physical cause-and-effect on top of their inherent reasoning abilities \cite{sanz2024reflective}. By fusing haptic and proprioceptive data with predictive world models, these agents better quantify uncertainty before acting \cite{wu2023daydreamer}. As illustrated in Figure \ref{figure2}, the resulting GPI architecture functions as adaptive hierarchical controller. This structure distinguishes between System-1 which is planning-oriented process that uses VLA models enhanced with haptic feedback to generate high-level strategies and System-2 is fast and reflexive controller as robotics foundation model that executes low-level actions based on high-level plans and real-time sensory feedback. This dual system approach bridges high-level reasoning with precise motor control to ensure generalization and robustness necessary for robots to handle diverse tasks and embodiments in dynamic industrial environments \cite{agarwal2025cosmos}. The performance of System-1, which is responsible for high-level strategy directly impacts the overall task outcome and is measured by the Success Rate (\%) and Generalization Score. In contrast, the execution fidelity of System-2, which governs precise motor control is quantified by the Pose Error (mm) and Angular Error (°), as numerically detailed out in Table 3. This dual-system approach bridges high-level reasoning with precise motor control to ensure generalization and robustness that is necessary for robots to handle diverse tasks and embodiments in dynamic industrial environments.\footnote{Deepmind Genie-3 is generative world model that creates interactive worlds by predicting next state from user actions and Genesis physics simulator calculates next state using deterministic solvers to test GPI agents in physically-plausible scenarios.}.

    \begin{figure*}[t]
       \centering
       \includegraphics[width=18cm]{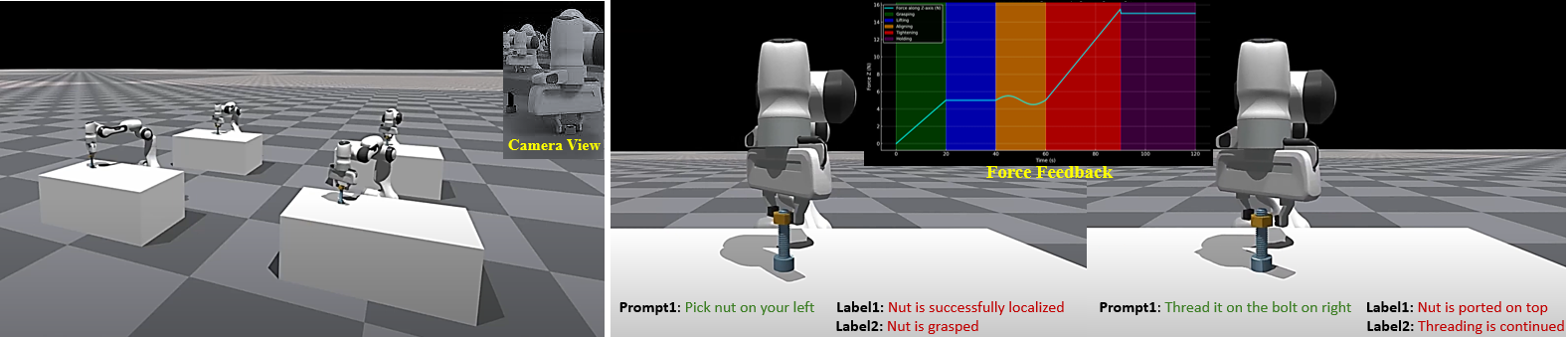}
       \caption{Franka Panda robot performing precision nut and bolt assembly using vision, language, haptic feedback and proprioception sensing for dynamic action planning in NVIDIA Isaac Sim.}
       \label{figure3}
       \vspace{-15pt}
    \end{figure*}

\subsection{Architectural Modifications for Industrial GPI}

To anchor utility of GPI framework, we outline architectural modifications aimed at evolving state-of-the-art VLA models into shared GPI-compatible interface for industrial tasks.

\begin{itemize}
    \item Gato: is adapted for GPI by expanding its token stream to include proprioceptive signals $(p_t)$, haptic feedback $(h_t)$ and embodiment descriptors $(\sigma)$ in addition to predicted future states $(\hat{s}_{t:t+H})$. This multimodal tokenization enables richer autoregressive sequence modeling that allows the model to encode object-action-effect affordance tuples. For example, in a multi-step assembly task, Gato predicts the consequences of a grasp and insertion by conditioning on both visual and tactile feedback thereby supporting long-horizon planning and robust transfer across robot arms and grippers. Technically, this requires integrating proprioceptive tokenizer e.g., MLP mapping joint states to tokens and a haptic encoder e.g., event-driven tactile arrays into its transformer input stream. \cite{liu2022robot}.
    \item RT-2: is updated by embedding proprioceptive and haptic feedback as additional tokens into the input sequence. Task-conditioned affordance priors are introduced via symbolic goal prompts and planning module is added that leverages dropout-based uncertainty estimation. For instance, in bin-picking scenario, RT-2 uses haptic feedback to detect slippage and replan its grasp while symbolic prompts specify the target object and placement location. The model is co-fine-tuned in both robotic trajectory data and Internet-scale vision-language tasks with actions and sensory feedback unified as text tokens \cite{Taylor2022}\cite{Nasiriany2025}. This enables RT-2 to generalize across platforms with variable dynamics, which is a key requirement for factories with diverse hardware.
    \item PaLM-E: natively supports multimodal input, but for GPI its encoder is extended to process unified streams of proprioception, vision, haptics, and embodiment descriptors. All inputs are represented as a sequence of tokens to allow the model to reason across heterogeneous data. For example, in quality inspection, PaLM-E can combine visual defect detection with force feedback to identify subtle assembly errors. The architecture uses pre-trained vision and language encoders with additional adapters for proprioceptive and haptic modalities and is trained on large-scale and multi-embodiment datasets \cite{Chen2025}\cite{Wang2024}. This supports complex manipulation in agile manufacturing.
    \item OpenVLA: is extended for GPI by infusing affordance-aware grounding and dynamic attention mechanisms. The model is modified to process task-relevant haptic and proprioceptive features, leveraging powerful visual encoders e.g., SigLIP, DINOv2 for fine-grained control. Task-prompted attention masking is incorporated to improve interpretability, which is critical for diagnostics and safety certification. For example, in industrial bin picking, OpenVLA can highlight which sensory features triggered failure by aiding root-cause analysis. Parameter-efficient prompt tuning e.g., learnable text prompts prepended to frozen encoders for rapid adaptation to new tasks \cite{Guruprasad2024}\cite{Wang2025}.
    \item VIMA: is extended for GPI by adding affordance conditioning, a spatial goal embedding module and a dynamic memory component. Curriculum learning and skill composition replay are introduced to facilitate transfer across related tasks such as sequential peg insertion and tool handovers. For example, VIMA can use cross-attention to focus on object-centric tokens thus enabling robust generalization to novel objects and manipulation sequences \cite{Lu2024}. The architecture supports multimodal prompt conditioning with object tokens and spatial embeddings that guide the autoregressive action decoder.
\end{itemize}

\subsubsection{Modality Fusion Architectures for Haptic Integration}

While visual and linguistic inputs provide rich semantic context but integrating haptic feedback is crucial for the physical grounding required in contact-rich manufacturing tasks. Fusing high-frequency continuous haptic data e.g., force, torque, tactile readings with lower-frequency visual data and discrete language tokens presents significant technical challenges. The asynchronous nature and differing dimensionalities of these modalities demand specialized architectural solutions within VLA models. A primary step involves processing raw haptic signals into format compatible with transformer-based architectures. This typically involves;

\begin{itemize}
    \item Rather than direct tokenization, which can lose nuanced information from continuous signals, haptic data streams are processed using dedicated encoders such as 1D Convolutional Neural Networks (CNNs), causal convolutions or Multi-Layer Perceptrons (MLPs). For example, a five-layer causal CNN can transform a window of 32 recent 6-axis force-torque readings into a fixed-size embedding that captures salient temporal features relevant to contact states or force profiles. Proprioceptive data e.g., joint positions, velocities are often encoded with MLPs for compact representation\cite{Schneider2023}.
    \item Handling temporal misalignment between, for instance a 1kHz force sensor stream and 30Hz camera feed is critical. Techniques include temporal pooling, interpolation or attention mechanisms that dynamically weigh information from different time steps across modalities. Some frameworks use self-supervised objectives to explicitly learn time-aligned representations thus improving model’s ability to capture cross-modal dependencies essential for manipulation \cite{Lygerakis2024}.
\end{itemize}

Once processed, these haptic embeddings must be integrated with visual and linguistic representations. Common fusion strategies include (i) A straightforward approach concatenates haptic, visual and language embeddings as part of input sequence to main transformer model \cite{Aziz2025}. While simple, this may not effectively capture complex inter-modal dependencies, (ii) More sophisticated methods employ cross-attention to allow tokens from one modality e.g., vision to attend to tokens from another e.g., haptics. Temporal cross-attention enables the model to dynamically weigh the importance of different senses based on task context \cite{Li2024}. For instance, during a precision peg-in-hole task, the model may initially rely on visual features for coarse alignment but shift attention to the haptic stream during final insertion to detect contact forces, resistance or successful seating, (iii) Some architectures fuse modality-specific embeddings into shared latent space using variational or self-supervised objectives or employ midst-mapping fusion where outputs of separate visual and haptic encoders are combined in final MLP for decision-making \cite{Schonfeld2019}. This approach is robust to missing or degraded modalities and supports generalization across tasks, (iv) Advanced frameworks introduce compensation branches or robust fusion modules to handle missing or low-quality haptic data to leverage shared features from visual modality to supplement or reconstruct haptic information to improve reliability in real-world industrial settings \cite{Yang2025}.

Forexample;  In peg insertion, model encodes visual and haptic streams with CNNs and causal convolutions fuses them via multimodal MLP and uses self-supervised objectives to align representations by enabling robust generalization to new geometries and perturbations \cite{lee2020making}. TVT-Transformer integrates tactile, visual and semantic text data using cross-modal self-attention by improving object recognition accuracy and adaptability in multimodal scenarios \cite{Chen2025a}. ForceVLA introduces a force-aware mixture-of-experts module that dynamically routes information between pretrained visual-language embeddings and real-time force feedback by enhancing dexterous manipulation in contact-rich tasks \cite{yu2025forcevla}. V-HCP employs upsampling and compensation modules to address modality imbalance and missing haptic data to achieve high success rates in flexible printed circuit assembly \cite{Jin2023}.

\subsection{Ablation Studies and Discussions}
To rigorously evaluate the grounding capabilities of our modified VLA-GPI models, we conducted series of comprehensive ablation studies. We selected two canonical high-precision manipulation tasks—nut-and-bolt insertion and dexterous timber-panel manipulation as these serve as robust benchmarks for assessing key attributes of industrial embodied intelligence including precise spatial reasoning, multimodal sensory integration and generalization under perturbations. The experiments were executed within NVIDIA Isaac Sim-5.2 environment that is chosen for its high-fidelity physics and rich sensor simulation in Figure \ref{figure3} and \ref{figure3}. To meet computational demands of multi-modal transformer-based inference and real-time simulation, all trials were performed on high-performance system i.e., Alienware Dell with 13th Gen Intel Core i9, 32 GB DDR5 RAM and NVIDIA RTX 4060\footnote{Experimental implementation of ablation studies presented can be accessed and reproduced using \url{https://drive.google.com/file/d/1-SvKeX1jOxSKNDcNd9gN0-J0o7v9x5xj/view?usp=drive_link}}.

\subsubsection{Nut-Bolt Assembly}
The experiment features 7-DoF Franka Emika FR3 robotic platform with parallel gripper, tasked with precision nut-and-bolt assembly within NVIDIA Isaac Sim environment, as shown in Figure \ref{figure3}. The objective is to insert nuts onto fixed bolts with tight thread-matching tolerance of 0.5 mm which is a task that demands precise alignment and compliant control. To ensure physical realism, high-fidelity CAD models of robot and components were used and CUda-optimized. The task rigorously tests for sub-millimeter positional accuracy, correct interpretation of simulated force-torque feedback and dynamic adaptation to variable starting positions via affordance modeling. To evaluate spatial generalization, the bolt target positions were randomized within 10 cm radius for each trial.

The quantitative results reported in Table \ref{tab:ablation-kpis} reveal differentiated capabilities of designated state of the art VLA-GPI models\footnote{Success Rate (\%) is percentage of trials completed successfully and reported as mean with standard deviation. Time (s) is average duration for successful trials. Pose Error (mm) is final Euclidean distance between end-effector and target pose by measuring translational accuracy. Angular Error ($^{\circ}$) is rotational difference by measuring orientation accuracy. Generalization Score is a normalized metric (0-1) quantifying performance consistency across randomized trials that is calculated as $1 - (\text{Standard Deviation of Success Rate} / \text{Mean Success Rate})$. A score closer to 1.0 indicates higher robustness to task variations}. Failure Modes categorizes primary reasons for task failure with counts in parentheses. Control Frequency (Hz) is rate at which the model sends commands to robot controller. A clear trade-off emerges amongst generalization, accuracy and inference speed. RT2-GPI emerged as top-performing model by achieving $93.3\%$ success rate and highest generalization score of $0.89$. Its superior performance is attributed to deeply integrated affordance modeling where fusion of vision, language, proprioceptive and tactile data enabled the model to infer correct alignment strategies even under significant spatial perturbations. Hence, this holistic understanding of object-centric possibilities and their physical outcomes reveal that;

\begin{itemize}
    \item VIMA-GPI showed high success of 88.3\% with effective geometric consistency due to object-centric memory. However, its moderate generalization score of $0.73$ suggests domain sensitivity stemming from affordances that are not deeply grounded into non-visual modalities.
    \item OpenVLA-GPI was notable for its fast average insertion time of 4.1 sec thus making it attractive for latency-sensitive applications. However, this speed came at the cost of precision of 1.12 mm pose error and robustness that result into lowest generalization score (0.67) that indicates a shallow integration of affordance cues.
\end{itemize}

In addition, Gato-GPI achieved respectable success rate of 83.3\% but was hampered by slowest inference time of 6.8 sec due to its autoregressive decoding mechanism. More decisively, its difficulty in adapting to novel configurations highlighted lack of explicit affordance encoding that reduces its flexibility in dynamic high-mix low-volume scenarios. For the agile manufacturing industry, where adaptability and precision are paramount, RT2-GPI and PaLM-E-GPI emerged as most balanced models that excell in precision, robustness and task adaptability. Their tight coupling of multimodal perception with sophisticated and affordance-aware reasoning enabled them to maintain high performance across significant domain shifts. This affirms their suitability for deployment in dynamic and precision-critical industrial scenarios where ability to reliably handle variation is key to operational success.

\begin{table*}[htbp]
\centering
\caption{Ablation Study on GPI-Extended Foundation Models for Nut-in-Bolt Task in NVIDIA Isaac Sim}  
\label{tab:ablation-kpis}
\scriptsize
\begin{tabularx}{\textwidth}{@{}l c c c c c c c@{}}
\toprule
\textbf{Model} & \textbf{Success Rate} & \textbf{Time} & \textbf{Pose Error} & \textbf{Angular Error} & \textbf{Generalization Score} & \textbf{Failure Modes} & \textbf{Control Frequency} \\
\midrule
\textbf{RT2-GPI}     & 93.3 $\pm$ 2.5 & 5.4 $\pm$ 0.15 & 0.53 $\pm$ 0.07 & 1.1 $\pm$ 0.15 & 0.89 & Jam (2), Misalignment (1) & 1.0 \\
\textbf{PaLM-E-GPI}  & 90.0 $\pm$ 3.0 & 5.9 $\pm$ 0.2 & 0.48 $\pm$ 0.055 & 1.0 $\pm$ 0.1 & 0.86 & Misalignment (2), Force spike (1) & 0.5 \\
\textbf{VIMA-GPI}    & 88.3 $\pm$ 2.0 & 5.7 $\pm$ 0.2 & 0.65 $\pm$ 0.09 & 1.3 $\pm$ 0.18 & 0.73 & Overshoot (3), Temporal lag (2) & 0.5 \\
\textbf{Gato-GPI}    & 83.3 $\pm$ 3.0 & 6.8 $\pm$ 0.35 & 0.87 $\pm$ 0.08 & 1.4 $\pm$ 0.2 & 0.71 & Slow convergence (3), Weak contact (4) & 1.0 \\
\textbf{OpenVLA-GPI} & 80.0 $\pm$ 3.5 & 4.1 $\pm$ 0.1 & 1.12 $\pm$ 0.095 & 1.7 $\pm$ 0.25 & 0.67 & Misalignment (4), Instability (3) & \text{0.5} \\
\bottomrule
\end{tabularx}
\end{table*}

\subsubsection{Dexterous Timber Panel Manipulation}

\begin{figure*}[t]
       \centering
       \includegraphics[width=18cm]{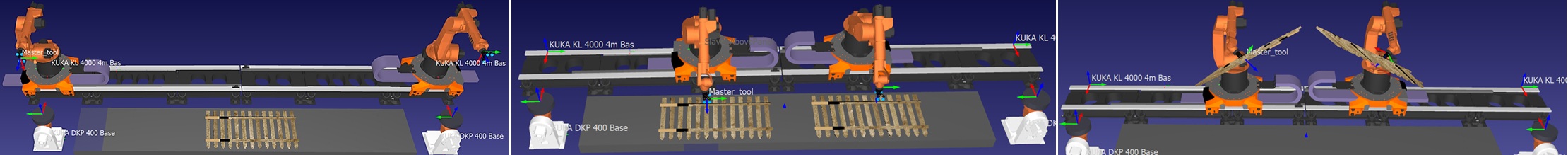}
       \caption{Dexterous timber panel manipulation task using KUKA manipulators mounted on linear tracks to perform coordinated grasping, lifting and reorientation of elongated timber panels. The scenario evaluates multimodal grounding, cooperative manipulation and spatial generalization under randomized panel and robot initial configurations.}
       \label{figure4}
       \vspace{-15pt}
    \end{figure*}

This experiment evaluates grounding and dexterous control capabilities of models through complex timber panel manipulation task performed within RoboDK suite. The setup (Figure \ref{figure4}) features two KUKA KR-series industrial manipulators mounted on linear tracks to enable large workspace coverage and coordinated motion. Each robot is paired with dual-axis positioner to allow precise orientation of the timber fixtures during cooperative assembly. The robots execute synchronized picking, reorientation and placement of elongated timber panels arranged on multi-slot fixture to simulate operations common in prefabricated construction and modular manufacturing pipelines. 

The task requires robots to grasp individual panels, lift them from the rack and realign them to predefined orientations with tight angular tolerances $\leq 1.5°$ while avoiding collisions with the fixtures, rails or partner robot. This introduces multi-modal grounding challenge involving coordinated bi-manual manipulation, reasoning over elongated deformable-like geometries and maintaining robust spatial consistency across extended trajectories. The elongated shape of the panels also amplifies errors in orientation estimation to make precise affordance interpretation critical for stable manipulation.

To assess robustness, panel positions on the rack were randomized within $±8 cm$ and their orientation jittered by ±5° for each trial. Additionally, robots were initialized from varied poses along the linear tracks thus requiring dynamic replanning and cross-arm coordination based on visual–geometric cues. This task probes several hallmark capabilities of grounded multimodal reasoning including (i) Geometric affordance inference for long slender objects with asymmetric mass distribution, (ii) Cooperative manipulation with implicit motion constraints between two agents, (iii) Predictive contact reasoning during grasping and reorientation, (iv) Generalization across significant spatial perturbations and viewpoint shifts. Models with stronger integration of spatial attention and affordance-aware reasoning i.e., RT2-GPI and PaLM-E-GPI have demonstrated smooth coordinated motion and consistent panel orientation accuracy. In contrast, those architectures that are relying heavily on auto-regressive decoding or primarily visual grounding exhibited difficulties with long-horizon consistency that resulted into drift, unstable grasp trajectories or poor adaptation to randomized initial poses.

\section{Challenges and Recommendations} \label{section4}

The pursuit of General Physical Intelligence (GPI) promises to transform automation by enabling robots to move beyond rigid and pre-programmed behaviors toward more adaptive systems capable of flexible perception, planning and interaction in diverse and unstructured environments. However, realizing industrially viable GPI through VLA models requires overcoming several foundational challenges including;

\subsubsection{Data Foundation for Physical Grounding} A major bottleneck for GPI is lack of comprehensive and physically grounded datasets. Most current large-scale datasets focus on vision and language but omit synchronized, high-frequency haptic (force-torque, tactile) and proprioceptive data streams that are essential for complex manufacturing tasks such as assembly, fastening and finishing. The sim2real gap is especially pronounced in contact-rich scenarios where simulations often fail to accurately model force exchanges, slippage and material deformation thus leading to policy failures during real-world deployment. To address these challenges, the following recommendations are proposed

\begin{itemize}
    \item There is an exigent need for benchmarks that provide synchronized streams of vision (RGBD), natural language instructions, end-effector states, robot kinematics and high-frequency haptic feedback. These datasets should also be annotated with affordance information to link sensory data to physical possibilities \cite{montesano2008learning}.
    \item Industrial viability requires more than task completion. Evaluation frameworks must evolve to include metrics that reflect real-world performance i.e., force-profile adherence for quality control, energy efficiency, compliant behavior and robustness to physical perturbations \cite{katyara2024benchmarking}.
    \item To train generalizable physical priors, new datasets should be generated in simulation with prime focus on contact dynamics. This involves extensive domain randomization of material properties i.e., stiffness, friction, surface roughness to ensure that learned policies are robust enough for the variability of real factories \cite{katyara2023data}.
\end{itemize}

\subsubsection{Fusion and Semantic Grounding of Haptic Feedback} Fusing asynchronous and high-dimensional haptic signals with dense visual information and symbolic language is a non-trivial task. Most extant models fail to leverage rich temporal dynamics and nuanced signals within haptic data that are essential for inferring critical information like contact state transitions, material compliance or physical intent \cite{wong2023vision}. Furthermore, semantically grounding linguistic commands like “press firmly” or “tighten until resistance” into precise and verifiable haptic strategies remains significant open problem for robotics community \cite{ding2024atom, katyara2021leveraging} to enable robots to achieve the level of dexterity required for many agile manufacturing tasks. Future work should prioritize following goals:
\begin{itemize}
    \item Develop multi-modal fusion architectures with temporal cross-attention mechanisms \cite{zhu2025advancing}. This would allow robot to dynamically weigh the importance of different senses based on task context for instance relying on vision to approach part and then shifting its focus to haptic feedback during final moments of precision insertion.
    \item Incorporate modality-specific encoders with learned priors over contact dynamics. This would enable language inputs to directly modulate robot physical behavior to allow it to translate abstract command like "handle delicately" into quantifiable control profile \cite{lee2020making}.
    \item Fine-tune pre-trained VLA models using interaction-rich episodes that are deeply grounded in haptic feedback. This will enhance their internal models of physical causality and tactile affordances to move beyond simple object recognition to deeper understanding of how objects behave under physical forces \cite{soyland2021grasping}.
\end{itemize} 

\subsubsection{Generalization and Sim-to-Real Transfer in Physical Interaction} Unlike perception or language tasks, physical interactions are heavily dependent on dynamic system parameters such as object mass, friction coefficients and material compliance \cite{li2025perception}. Generalizing policies across such variations, a daily reality in manufacturing where component suppliers or material batches change, has yet to be fully addressed. This problem is compounded by sim2real gap which is most pronounced in contact dynamics. Inaccuracies in simulating force exchange, slippage or deformation of non-rigid parts cause policies trained in simulation to fail catastrophically~\cite{li2024large}. Overcoming this requires creating more robust and adaptive policies. Future research should prioritize;

\begin{itemize}
    \item Move beyond generic randomization to more targeted approaches. This involves focusing simulation on specific range of physical properties that robot is likely to encounter during industrial deployments to encourage learning of tailored and effective resilient policies \cite{chen2023multimodality}.
    \item Integrate meta-learning and zero-shot generalization capabilities. The goal is to develop VLA-haptic systems that can quickly adapt to new components or tasks with minimal real-world data that drastically reduce time and cost associated with production line changeovers \cite{bucher2021handling}.
\end{itemize}
   
\subsubsection{Hierarchical Planning and Feedback-Driven Control}

 Complex manufacturing workflows demand long-horizon planning where abstract goals must be decomposed into sequence of executable sub-tasks. Coordinating high-level plan often generated by VLA model with low-level controller to adapt to contact conditions in real time introduces significant layered synchronization problems. Furthermore, reliably handling execution failures, unexpected contact events and task transitions based on force thresholds requires resilient integration of feedback into controller\footnote{Crux is not synchronization between sequential planning layers but mismatch in temporal scales and uncertainty propagation in hierarchical stack.}. Future research focus to;
\begin{itemize}
    \item Develop hierarchical control architectures with bidirectional communication pathways. In such system, the high-level planner does not simply issue commands but the low-level controller also reports back critical information. For example, sensory feedback indicating slippage, unexpected resistance or force discontinuities would inform mid-level decision modules to adapt current strategy or alert high-level planner to flag anomaly such as part defect that requires completely new motion plan \cite{ding2023task}.
    \item Automate learning of reusable temporal and functional task structures from demonstrations. By identifying and encoding fundamental skills i.e., "pick" ,"align", "insert", policies are constructed more efficiently and composed into more complex and long-horizon tasks. This encourages skill reuse across different production lines and speeds up process of programming new tasks \cite{tavassoli2023learning}.
\end{itemize}

\subsubsection{Real-Time Constraints, Interpretability and Safety} When scaled to billions of parameters, VLA models often fail to meet strict real-time demands of industrial control loops that frequently operate at high frequencies. Furthermore, their "black box" nature lacks interpretability that is required for rigorous failure analysis, safety validation and regulatory certification. This opacity is significant barrier to adoption in high-stakes manufacturing environments where system predictability is preeminent. Therefore, we recommend; 
\begin{itemize}
    \item Pursue aggressive model compression, distillation and quantization to create lightweight versions of GPI systems that can be deployed on cost-effective edge devices \cite{polino2018model}. Utilizing real-time operating systems (RT-PREEMPT patches) is essential to guarantee microsecond determinism required for safe physical interaction.
    \item Incorporate Explainable AI (XAI) techniques specifically tailored to physically interactive systems. Instead of generic approaches, this involves implementing concrete methods such as attention visualization to identify which specific visual or haptic cues the model can use to make critical decisions on executing force-threshold command. This is invaluable for post-hoc failure analysis. Furthermore, integrating methods like conformal prediction layers can provide statistically rigorous confidence bounds for action predictions by enabling system to anticipate failures and proactively request human assistance when uncertainty exceeds safety threshold. The goal is to create auditable models that link their actions directly to specific perceptual and haptic triggers to provide transparency to satisfy safety standards and regulatory certifications \cite{gorriz2023computational}.
\end{itemize}

\subsubsection{Role of Generative AI in Data Augmentation and System Design}

Generative AI (GenAI) is rapidly transforming the landscape of robotics and manufacturing by addressing "contact-rich data gap" and enabling co-design of intelligent systems and environments. Recent research demonstrates that GenAI is not only a tool for data augmentation but also a catalyst for next generation of cyber-physical manufacturing systems.

GenAI leverages deep generative models such as diffusion models, GANs, VAEs and autoregressive models to synthesize large-scale high-fidelity datasets for contact-rich manipulation and multi-modal robotic learning \cite{Yang2025a}\cite{Urain2024}. Physics-driven pipelines now integrate human demonstrations, trajectory optimization and simulation to generate physically consistent and cross-embodiment datasets to enable zero-shot policy transfer to real hardware with minimal human input \cite{Hsu2025}\cite{BondTaylor2021}. Tools like RoboGen and RoboManuGen automate generation of diverse tasks, scenes and training supervisions by scaling up skill acquisition and anomaly detection in manufacturing robots \cite{wang2023robogen}\cite{Xian2023}. Diffusion-based models such as M2Diffuser directly generate whole-body motion trajectories for mobile manipulation thereby incorporating physical constraints and energy functions to ensure robust and safe execution in real-world environments \cite{Yan2025}\cite{Li2025}.

GenAI extends beyond data to generative design of manufacturing systems. Cognitive digital twins and generative planning software integrated with IoRT and cyber-physical systems enable the simulation and optimization of workcell layouts, robot embodiments and process flows in immersive industrial metaverse environments \cite{Liu2019}\cite{Mata2025}. Diffusion and large language models can propose novel and physics-plausible workcell configurations, simulate complex multi-agent interactions and optimize for criteria such as collision-free motion, energy efficiency and adaptability to high-mix manufacturing \cite{Daiya2025}\cite{Mikolajewska2025}. This generative co-design paradigm accelerates transition from Industry 4.0 to Industry 5.0 where AI not only learns policies but also iteratively refines the physical and digital infrastructure of smart factories.

\subsubsection{Resilience through Decentralized and Modular Architectures}

Ensuring resilience is a central challenge in deploying GPI for agile manufacturing. Resilience is defined as system ability to maintain or rapidly recover stable productivity in the face of major disruptions or persistent stresses. Recent research demonstrates that resilience is best achieved through decentralized, modular and multi-agent system (MAS) architectures that fundamentally enhance fault tolerance, adaptability and scalability compared to monolithic and centrally controlled production lines.
Modular production systems enabled by MAS allow for dynamic reconfiguration and distributed decision-making. Each module or agent often encapsulated as an Asset Administration Shell (AAS) possesses self-description, self-configuration and self-optimization capabilities to support plug-and-produce functionality and rapid adaptation to changing production demands or failures \cite{Sidorenko2023}\cite{Bi2024}. In such architectures, if one agent or module fails others can autonomously reallocate tasks and maintain workflow continuity thereby minimizing single points of failure and enabling graceful degradation under resource constraints. 
Decentralized control is realized through peer-to-peer negotiation, distributed intelligence and local autonomy. Advanced frameworks such as blockchained smart contract pyramid-driven multi-agent autonomous process control (BSCP-MAAPC) and digital twins (e.g., ManuChain II) use blockchain-secured smart contracts to coordinate task allocation, schedule adjustments and process verification without reliance on a central authority \cite{Khan2005}\cite{Leng2023}. These systems enable verifiable, tamper-proof and transparent operations to support rapid dynamic adjustment to disruptions and continuous improvement via decentralized deep learning.
Integrating edge-AI trust engines and consensus mechanisms further enhances resilience by enabling real-time anomaly detection, predictive maintenance and trust scoring among heterogeneous agents even in adversarial or uncertain environments \cite{Huang2021}\cite{Ucar2024}. This distributed intelligence allows for proactive task reassignment and robust operation despite network or agent failures.

\section{Conclusion} \label{section5}

This practical review has systematically surveyed state-of-the-art of Vision-Language-Action (VLA) models as foundational technology for achieving General Physical Intelligence (GPI) in the context of resilient and agile manufacturing. Our goal was to assess the current readiness of these powerful models for industrial deployment and to identify critical gaps that remain. Through comprehensive literature review organized into  thematic pillars including multisensory representation learning, sim2real transfer, planning and control, uncertainty and safety and benchmarking, we have highlighted significant advancements alongside persistent challenges.

Our ablation studies on GPI-enhanced VLA models that are exemplified by strong performance of RT2-GPI in precision assembly task underscores potential of these architectures when augmented with affordance modeling and multimodal sensory feedback. However, the study also revealed clear trade-offs between generalization, accuracy and inference speed across different models which is indicating that no single current framework offers a complete solution. The path toward industrially viable GPI systems requires addressing fundamental obstacles as detailed in our analysis. The critical challenges include bridging the contact-rich data gap, developing robust techniques for fusing and semantically grounding haptic feedback to overcome persistent sim2real gap in physical interaction dynamics, ensure safety and interpretability under strict real-time constraints and build system-level resilience. Addressing these challenges through focused research including development of better datasets, advanced fusion architectures, targeted simulation, hierarchical control and physically grounded XAI will be crucial. Despite these hurdles, the rapid progress in foundation models offers promising trajectory toward creating truly intelligent, adaptable and dependable robotic partners capable of realizing human-centric and resilient automation goals of Industry 5.0.

\bibliographystyle{IEEEtran}
\bibliography{GPI_paper_biblography_updated}

\end{document}